\documentclass{article} 
\usepackage{iclr2020_conference,times}

\usepackage[utf8]{inputenc} 
\usepackage[T1]{fontenc}    
\usepackage{hyperref}       
\usepackage{url}            
\usepackage{booktabs}       
\usepackage{amsfonts}       
\usepackage{nicefrac}       
\usepackage{microtype}      
\usepackage{graphicx}
\usepackage{subfigure}
\usepackage{amsmath}
\usepackage{amssymb}
\usepackage{bm}
\usepackage{makecell}
\usepackage{comment}
\usepackage{hyperref}
\usepackage{todonotes}
\usepackage{wrapfig}
\usepackage{float}

\newcommand{\mbf}[1] {\mathbf{#1}}

\newcommand{\vv}{\mathbf{v}}
\newcommand{\pp}{\mathbf{p}}
\newcommand{\cc}{\mathbf{c}}
\newcommand{\mm}{\mathbf{m}}

\newcommand{\cut}[1]{}

\title{Physics-as-Inverse-Graphics: \\ Unsupervised Physical Parameter \\Estimation from Video}

\author{%
  Miguel Jaques \\
  School of Informatics\\
  University of Edinburgh\\
  Edinburgh, UK \\
  \texttt{m.a.m.jaques@sms.ed.ac.uk} 
    \And
    Michael Burke \\
  School of Informatics\\
  University of Edinburgh\\
  Edinburgh, UK \\
  \texttt{michael.burke@ed.ac.uk} 
  \And
  Timothy Hospedales \\
  School of Informatics \\
  University of Edinburgh \\
  Edinburgh, UK \\
  \texttt{t.hospedales@ed.ac.uk} 
}

\iclrfinalcopy 
\begin{document}

\maketitle

\begin{abstract}
We propose a model that is able to perform unsupervised physical parameter estimation of systems from video, where the differential equations governing the scene dynamics are known, but labeled states or objects are not available. Existing physical scene understanding methods require either object state supervision, or do not integrate with differentiable physics to learn interpretable system parameters and states. We address this problem through a \textit{physics-as-inverse-graphics} approach that brings together vision-as-inverse-graphics and differentiable physics engines, enabling objects and explicit state and velocity representations to be discovered. This framework allows us to perform long term extrapolative video prediction, as well as vision-based model-predictive control. Our approach significantly outperforms related unsupervised methods in long-term future frame prediction of systems with interacting objects (such as ball-spring or 3-body gravitational systems), due to its ability to build dynamics into the model as an inductive bias. We further show the value of this tight vision-physics integration by demonstrating data-efficient learning of vision-actuated model-based control for a pendulum system. We also show that the controller's interpretability provides unique capabilities in goal-driven control and physical reasoning for zero-data adaptation. 

\end{abstract}

\section{Introduction}
System identification or physical parameter estimation is commonly required for control or state estimation for physical modelling, and typically relies on dedicated sensing equipment and carefully constructed experiments.
Current machine learning approaches to physical modeling from video either require training by supervised regression from video to object coordinates before estimating explicit physics \citep{Watters2017VisualVideo,Wu2017LearningDe-animation,Belbute-Peres2018End-to-EndControl}, or are able to discover and segment objects from video in an unsupervised manner, but do not naturally integrate with a physics engine for long-term predictions or  generation of interpretable locations and physical parameters for physical reasoning  \citep{Xu2019UnsupervisedDynamics,vanSteenkiste2018RelationalInteractions}. In this work, we bridge the gap between unsupervised discovery of objects from video and learning the physical dynamics of a system, by learning unknown physical parameters and explicit trajectory coordinates. \cut{This opens the door to long term video prediction; applications in vision-actuated model-based control, where we have access to video streams but not the underlying object states; and even counterfactual physical reasoning.}


Our approach, called \textit{physics-as-inverse-graphics}, solves the physical modeling problem via a novel vision-as-inverse-graphics encoder-decoder system that can render and de-render image components using Spatial Transformers (ST) \citep{Jaderberg} in a way that makes it possible for the latent representation to generate disentangled interpretable states (position/velocity). These can be used directly by a differentiable physics engine \citep{Degrave2016ARobotics,Belbute-Peres2018End-to-EndControl} to learn the parameters of a scene where the family of differential equations governing the system are known (e.g. objects connected by a spring), but the corresponding parameters are not (e.g.  spring constant). \cut{This way, physics-as-inverse graphics enables the incorporation of high-level physical interaction knowledge into the learning process as an inductive bias.} This allows us to to identify physical parameters and learn vision components of the model jointly in an end-to-end fashion. Our contribution is a solution to unsupervised learning of physical parameters from video, without having access to ground-truth appearance, position or velocities of the objects, a task that had so far remained unsolved \citep{Wu2015GalileoLearning,Belbute-Peres2018End-to-EndControl}.

In addition to showing that our model can learn physical parameters without object or state supervision (a task with intrinsic scientific interest in and of itself), we show that incorporating dynamics priors in the form of known physical equations of motion with learnable parameters together with learnable vision and graphics can improve model performance in two challenging tasks: long term video prediction and visual model predictive control. We first evaluate physical parameter estimation accuracy and future video frame prediction on 4 datasets with different non-linear interactions and  visual difficulty. We then demonstrate the value of our method by applying it for data-efficient learning of vision-based control of an under-actuated pendulum. Notably our unique ability to extract interpretable states and parameters from pixels without supervision enables end-to-end vision-based control to exploit goal-paramaterized policies and physical reasoning for zero-shot adaptation.



\section{Related Work}

The ability to build inductive bias into models through structure is a key factor behind the success of modern neural architectures. Convolutional operations capture spatial correlations \citep{fukushima1980neocognitron} in images, recurrency allows for temporal reasoning \citep{hochreiter1997long}, and spatial transformers \citep{Jaderberg} provide spatial invariance in learning. However, many aspects of common data generation processes are not yet considered by these simple inductive biases. Importantly, they typically ignore the physical interactions underpinning data generation. For example, it is often the case that the underlying physics of a dynamic visual scene is known, even if specific parameters and objects are not. Incorporation of this information would be beneficial for learning, predicting the future of the visual scene, or control. Physics-as-inverse graphics introduces a framework that allows such high-level physical interaction knowledge  to be incorporated into learning, even when ground-truth object appearance, positions and velocities are not available.

In recent years there has been increased interest in physical scene understanding from video \citep{Fragkiadaki2016LearningBilliards, Finn2016UnsupervisedPrediction, Fraccaro2017ALearning, Chang2017ADynamics, Jonschkowski2017PVEs:Representations, Zheng2018UnsupervisedNetworks, Janner2019ReasoningPlanning}.  
In order to learn explicit physical dynamics from video our system must discover and model the objects in a scene, having position as an explicit latent variable. Here we build on the long literature of neural vision-as-inverse-graphics \citep{Hinton, Kulkarni, Huang, Ellis2018LearningImages,  Romaszko2017Vision-as-Inverse-Graphics:Image,  Wu2017NeuralDe-rendering}, particularly on the use of spatial transformers (ST) for rendering \citep{Eslami,Rezende2016One-ShotModels,Zhu2018Object-OrientedPredictor}.

There are several models that assume knowledge of the family of equations governing system dynamics, but where the individual objects are either pre-segmented or their ground-truth positions/velocities are known \citep{Stewart2017Label-FreeKnowledge,Wu2017LearningDe-animation,Belbute-Peres2018End-to-EndControl}. In terms of learning physical parameters, our work is directly inspired by the Galileo model and Physics 101 dataset \citep{Wu2015GalileoLearning,Wu2016PhysicsVideos}, which fits the dynamics equations to a scene with interacting objects. However, the Galileo model makes use of custom trackers which estimate the position and velocity of each object of interest, and is incapable of end-to-end learning from video, thus bypasses the difficulty of recognizing and tracking objects from video using a neural system. To the best of our knowledge, our model is the first to offer end-to-end unsupervised physical parameter and state estimation. 

Within the differentiable physics literature \citep{Degrave2016ARobotics},  \citet{Belbute-Peres2018End-to-EndControl} observed that a multi-layer perceptron (MLP) encoder-decoder architecture with a physics engine was not able to learn without supervising the physics engine's output with position/velocity labels (c.f. Fig.~4 in \cite{Belbute-Peres2018End-to-EndControl}). While in their case 2\% labeled data is enough to allow learning, the transition to \textit{no} labels causes the model to not learn at all. \cut{The difficulty of incorporating deterministic physics engines into unsupervised learning models has prohibited the exploitation of this form of inductive bias.} The key contribution of our work is the incorporation of vision-as-inverse-graphics with physics, which makes the transition possible.

Another related area of increasing interest is unsupervised discovery of objects and/or dynamics from video \citep{Xu2019UnsupervisedDynamics,vanSteenkiste2018RelationalInteractions,Greff2019Multi-ObjectInference,Burgess2019MONet:Representation}. Though powerful, such models do not typically use interpretable latent representations that can be directly used by a physics engine, reasoned about for physical problem solving, or that are of explicit interest to model users. For example, \citet{Kosiorek2018SequentialObjects} and  \citet{Hsieh2018LearningPrediction} use ST's to locate/place objects in a scene and predict their motion, but this work differs from ours in that our coordinate-consistent design obtains explicit cartesian, angular or scale coordinates, allowing us to feed state vectors directly into a differentiable physics engine. Under a similar motivation as our work, but without an inverse-graphics approach, \citet{Ehrhardt2018UnsupervisedObservations} developed an unsupervised model to obtain consistent object locations. However, this only applies to cartesian coordinates, not angles or scale.



Despite recent interest in model-free reinforcement learning, model-based control systems have repeatedly shown to be more robust and sample efficient \citep{deisenroth2011pilco,ManiaRech18,Watters2019COBRA:Exploration}. \citet{hafner2018Planet} learn a latent dynamics model (PlaNet) that allows for planning from pixels, which is significantly more sample efficient than model-free learning strategies A3C \citep{mnih2016asynchronous} and D4PG \citep{barth2018distributed}. However, when used for control, there is often a desire for visually grounded controllers operating under known dynamics, as these are  verifiable and interpretable \citep{Burke19Explanation}, and provide transferability and generality. However, system identification is challenging in vision-based control settings. \citet{Byravan2018SE3-Pose-Nets:Control} use supervised learning to segment objects, controlling these using known rigid body dynamics. \citet{penkov2019learning} learn feedforward models with REINFORCE \citep{Williams1992} to predict physical states used by a known controller and dynamical model, but this is extremely sample inefficient. In contrast, we learn parameter and state estimation modules jointly to perform unsupervised system identification from pixels, enabling data-efficient vision-actuated model-based control.

\begin{figure*}[t]
  \centering
    \includegraphics[width=0.9\textwidth]{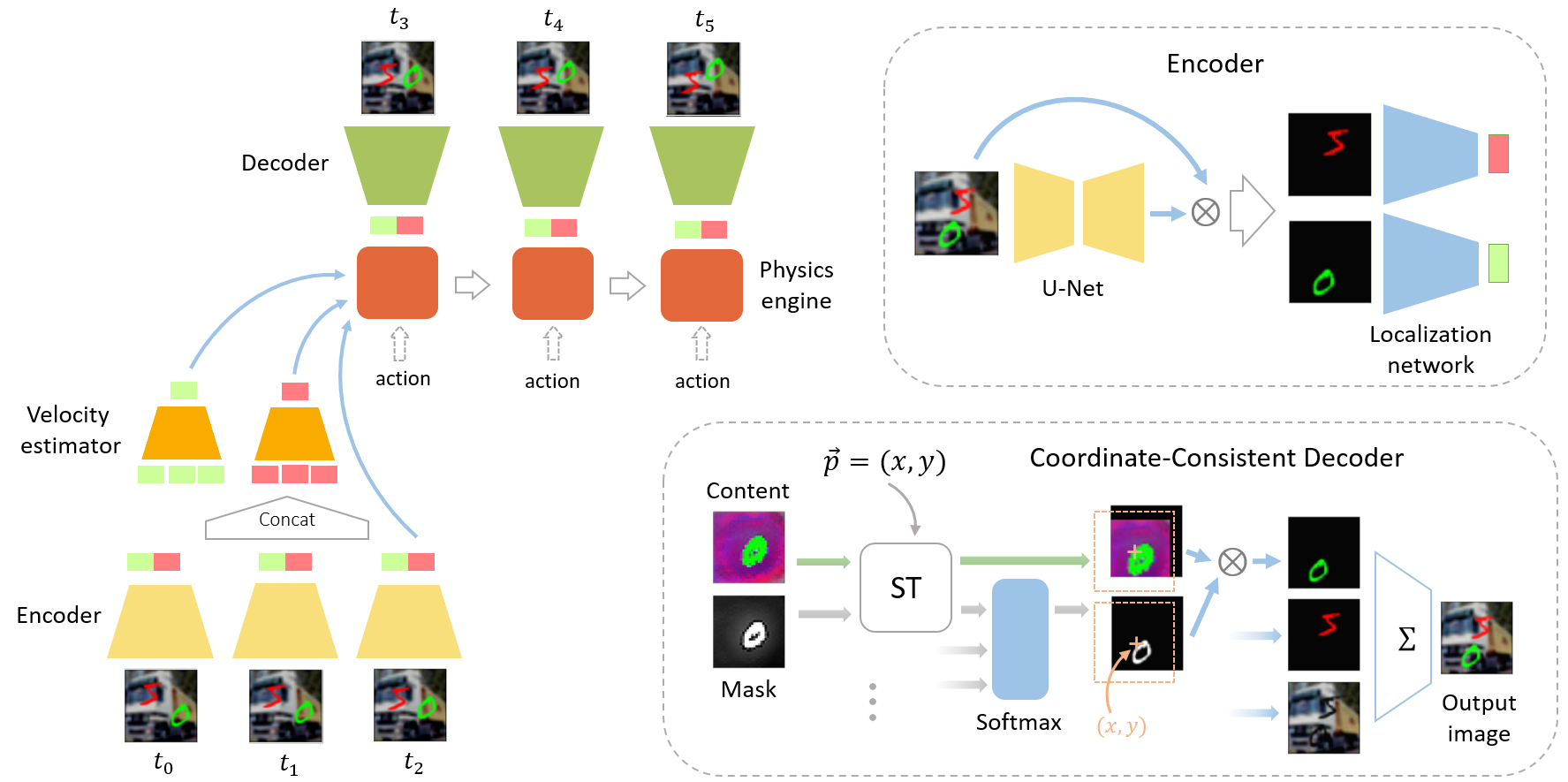}
  \caption{\textbf{Left:} High-level view of our architecture. The encoder (\textbf{top-right}) estimates the position of $N$ objects in each input frame. These are passed to the velocity  estimator which estimates objects' velocities at the last input frame. The positions and velocities of the last input frame are passed as initial conditions to the physics engine. At every time-step, the physics engine outputs a set of positions, which are used by the decoder (\textbf{bottom-right}) to output a predicted image. If the system is actuated, an input action is passed to the physics engine at every time-step. See Section 3 for detailed descriptions of the encoder and decoder architectures.}
  \label{fig:model-overview}
\end{figure*}
\section{Learning Physical Parameters from Video via Inverse Graphics}


In order to learn explicit physics from video, several components have to be in place. First, the model must be able to learn to identify and represent the objects in an image. In order to perform dynamics prediction with a physics engine, the position and velocity of the object must be represented as explicit latent states (whereas appearance can be represented through some latent vector or, in our case, as a set of learned object templates).
Our sequence-to-sequence video prediction architecture consists of 4 modules trained jointly: an encoder, a velocity estimator, a differentiable physics engine, and a graphics decoder. The architecture is shown in Figure~\ref{fig:model-overview}.

    
\textbf{Encoder} {} The encoder net takes a single frame $I_t$ as input and outputs a vector $\pp_t \in \mathbb{R}^{N\times D}$ corresponding to the $D$-dimensional coordinates of each of $N$ objects in the scene, $\pp_t = [\pp_t^1,...,\pp_t^N]$. For example, when modelling position in 2D space we have $D=2$ and $\pp_t^n = [x,y]^n_t$; when modelling object angle we have $D=1$ and $\pp_t^n = [\theta_t^n]$. The encoder architecture is shown in Figure~\ref{fig:model-overview}(top right).


To extract each object's coordinates we use a 2-stage localization approach\footnote{Though any other architecture capable of effectively extracting object locations from images would work.}. First, the input frame is passed through a U-Net \citep{Ronneberger2015U-Net:Segmentation}  to produce $N$ unnormalized masks. These  masks (plus a learnable background mask) are stacked and passed through a softmax to produce $N+1$ masks, where each input pixel is softly assigned to a mask. The input image is then multiplied by each mask, and a 2-layer location network produces coordinate outputs from each masked
input component. 
For a 2D system where the coordinates of each object are its $(x,y)$ position (the polar coordinates case is analogous) and the images have dimensions $H \times H$, the encoder output represents $(x,y)$ coordinates with values in $[0,H]$. To do this, the activation of the encoder's output layer is a saturating non-linearity $H/2\cdot tanh(\cdot)+ H/2$.

\textbf{Velocity estimator} {} The velocity estimator computes the velocity vector of each object at the $L$-th input frame given the coordinates produced by the encoder for this object at the first $L$ input frames, $\vv_L^n = f(\pp_1^n, ..., \pp_L^n)$. We implement this as a 3 hidden layer MLP with 100 tanh activated units. 

\textbf{Differentiable physics engine} {} The physics engine contains the differential equations governing the system, with unknown physical parameters to be learned -- such as spring constants, gravity, mass, etc. Given initial positions and velocities produced by the encoder and velocity estimator, the physics engine rolls out the objects' trajectories. In this work we use
a simple physics engine with Euler integration, where $\pp_t, \vv_t$ is computed from $\pp_{t-1}, \vv_{t-1}$ by repeating for $i\in [1..M]$:
\begin{equation}
    \pp_{t+\frac{i}{M}} = \pp_{t+\frac{i-1}{M}} + \frac{\Delta t}{M} \cdot \vv_{t+\frac{i}{M}} \quad ; \quad \vv_{t+\frac{i}{M}} = \vv_{t+\frac{i-1}{M}} + \frac{\Delta t}{M} \cdot \mbf{F}(\pp_{t+\frac{i-1}{M}},\vv_{t+\frac{i-1}{M}}; \theta) \,,
\end{equation}
where $\Delta t$ is the integration step, $\theta$ are the physical parameters and $\mbf{F}$ is the force applied to each object, according to the equations in Appendix A. We use $M=5$ in all experiments. In principle, more complex physics engines could be used \citep{Chen2018NeuralEquations,Belbute-Peres2018End-to-EndControl}.


\textbf{Coordinate-Consistent Decoder} {} The decoder takes as input the positions given by the encoder or physics engine, and outputs a predicted image $\tilde{I}_t$.
The decoder is the most critical part of this system, and is what allows the encoder, velocity estimator and physics engine to train correctly in a fully unsupervised manner. We therefore describe its design and motivation in greater detail.

 While an encoder with outputs in the range $[0, H]$ \textit{can} represent coordinates in pixel space, it does not mean that the decoder \textit{will} learn to correctly associate an input vector $(x,y)$ with an object located at  pixel  $(x,y)$. If the decoder is  unconstrained, like a standard MLP, it can very easily learn erroneous, non-linear representations of this Cartesian space. For example, given two different inputs, $(x_1,y_1)$ and $(x_1, y_2)$, with $y_1\neq y_2$, the decoder may render those two objects at different horizontal positions in the image. While having a correct Cartesian coordinate representation is not strictly necessary to allow physical parameters of the physics engine to be learned from video, it is critical to ensure correct future predictions. This is because the relationship between position vector and pixel space position must be fixed: if the position vector changes by $(\Delta x, \Delta y)$, the object's position in the output image must change by $(\Delta x, \Delta y)$. This is the key concept that allows us to improve on \cite{Belbute-Peres2018End-to-EndControl},  in order to learn an encoder, decoder and physics engine without state labels.



In order to impose a correct latent-coordinate to pixel-coordinate correspondence, we use spatial transformers (ST) with inverse parameters as the decoder's writing attention mechanism. We want transformer parameters $\omega$ to be such that a decoder input of $\pp_t^n = [x,y]_t^n$, places the center of the writing attention window at $(x,y)$ in the image, or that a decoder input of $\pp_t^n = \theta_t^n$ rotates the attention window by $\theta$. 
In the original ST formulation \citep{Jaderberg}, the matrix $\omega$ represents the affine transformation applied to the \textit{output} image to obtain the \textit{source} image. This means that the elements of $\omega$ in Eq.~1 of \citet{Jaderberg} do not directly represent translation, scale or angle of the writing attention window. To achieve this representation, we use a ST with inverse transformation parameters. For a general affine transformation with translation $(x,y)$, angle $\theta$ and scale $s$, we want to modify the source image coordinates according to:
\begin{equation}
    \begin{pmatrix} 
    x_o \\
    y_o \\
    1
    \end{pmatrix}
    =
    \begin{pmatrix} 
    s\cdot \cos \theta & s\cdot \sin \theta & x\\
    -s\cdot \sin \theta & s\cdot \cos \theta & y\\
    0 & 0 & 1
    \end{pmatrix}
    \begin{pmatrix} 
    x_s \\
    y_s \\
    1
    \end{pmatrix}
\label{eq:transf}
\end{equation}

This transformation can be obtained with a ST by inverting \eqref{eq:transf}:
\begin{equation}
    \begin{pmatrix} 
    x_s \\
    y_s \\
    1
    \end{pmatrix}
    = \frac{1}{s}
    \begin{pmatrix} 
    \cos \theta & -\sin \theta & -x\cos{\theta} + y \sin{\theta} \\
    \sin \theta & \cos \theta & -x\sin{\theta}  -y\cos{\theta} \\
    0 & 0 & s
    \end{pmatrix}
    \begin{pmatrix} 
    x_o \\
    y_o \\
    1
    \end{pmatrix}
\label{eq:inv_transf}
\end{equation}
Therefore, to obtain a decoder with coordinate-consistent outputs, we simply use a ST with parameters $\omega$ as given in \eqref{eq:inv_transf}

\cut{\footnote{In a Pytorch/Tensorflow spatial transformer implementation, the translation parameters actually correspond to the number of pixels as fraction of the image size, and the difference in resolution between the source and output has to be accounted for. This means that, in practice, instead of $-x\cos\theta+y\sin\theta$, the code uses  $2\cdot((H-x)\cos\theta-(H-y)\sin\theta)/H$.}.}



Each object is represented by a learnable content $\cc^n \in [0,1]^{H\times H \times C}$ and mask $\mm^n \in \mathbb{R}^{H\times H \times 1}$ tensor, $n=1..N$. Additionally, we learn background content $\cc^{bkg} \in [0,1]^{H\times H \times C}$ and  mask $\mm^{bkg} \in \mathbb{R}^{H\times H \times 1}$, that do not undergo spatial transformation. One may think of the content as an RGB image containing the texture of an object and the mask as a grayscale image containing the shape and z-order of the object. In order to produce an output image, the content and mask are transformed according to $[\hat{\cc}_t^n, \hat{\mm}_t^n] = \text{ST}([\cc^n, \mm^n], \omega_{\pp_t^n})$ 
and the resulting logit masks are combined via a softmax across channels, $[\tilde{\mm}_t^1,..., \tilde{\mm}_t^N, \tilde{\mm}_t^{bkg}] = \text{softmax}(\hat{\mm}_t^1,..., \hat{\mm}_t^N, \mm^{bkg})$.
The output image is obtained by multiplying the output masks by the contents:
\begin{equation}
\tilde{I}_t = \tilde{\mm}_t^{bkg} \odot \cc^{bkg} + \sum_{n=1}^N \tilde{\mm}_t^n \odot \hat{\cc}_t^n.
\end{equation}
The decoder architecture is shown in Fig. \ref{fig:model-overview}, bottom-right. The combined use of ST's and masks provides a natural way to model depth ordering, allowing us to capture occlusions between objects.



\begin{figure*}[t]
  \centering
    \includegraphics[width=0.85\textwidth]{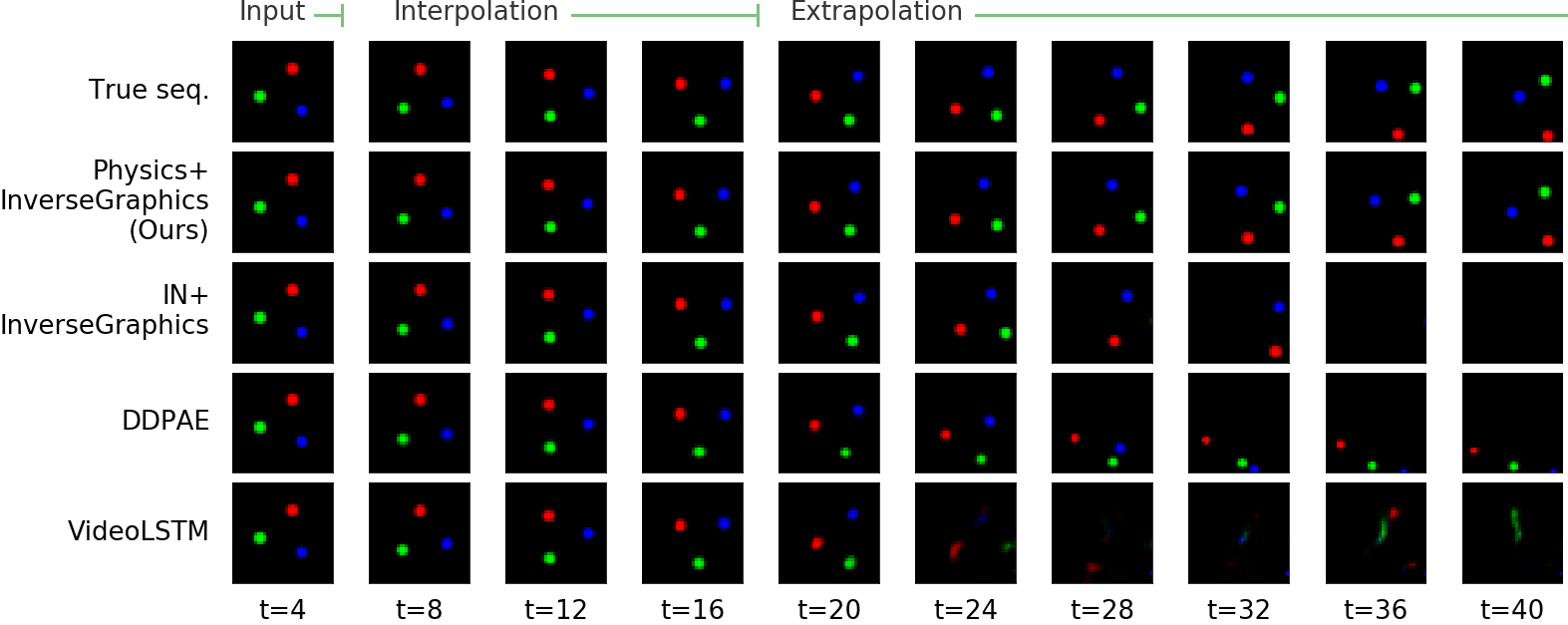}
    
    \vspace{0.25cm}
    \includegraphics[width=0.85\textwidth]{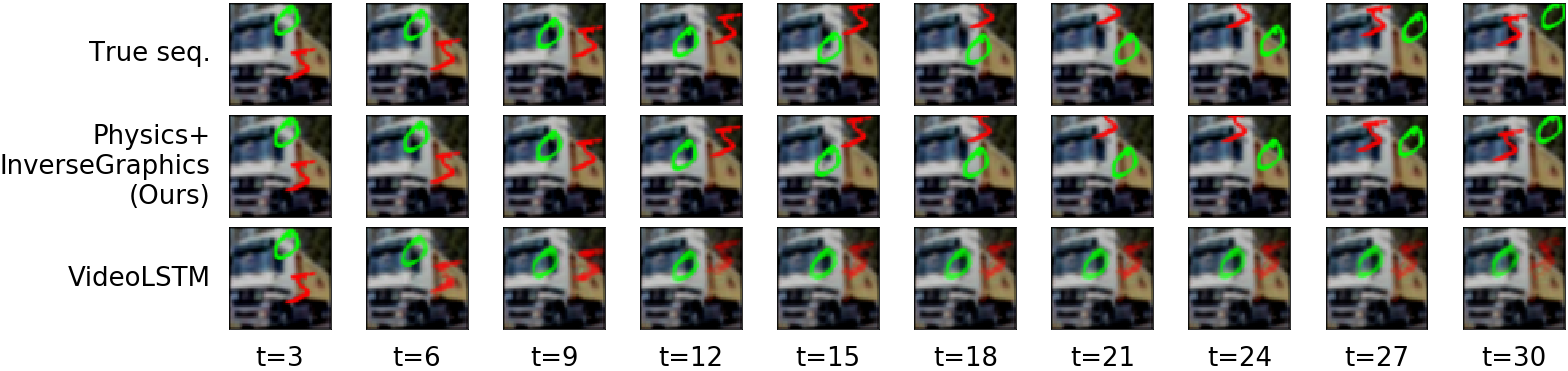}
    \vspace{-0.1cm}
  \caption{Future frame predictions for 3-ball gravitational system (\textbf{top}) and 2-digit spring system (\textbf{bottom}). 
  IN: Interaction Network. Only the combination of Physics and Inverse-Graphics maintains object integrity and correct dynamics many steps into the future.}
  \label{fig:predictions}
\end{figure*}
\vspace{-0.1cm}

\textbf{Auxiliary autoencoder loss} {}
Using a constrained decoder ensures the encoder and decoder produces objects in consistent locations. However, it is hard to learn the full model from future frame prediction alone, since the encoder's training signal comes exclusively from the physics engine.
To alleviate this and quickly build a good encoder/decoder representation, we add a static per-frame autoencoder loss. 


\textbf{Training} {} During training we use $L$ input frames and predict the next $T_{pred}$ frames. Defining the frames produced by the decoder via the physics engine as $\tilde{I}_t^{\text{pred}}$ and the frames produced by the decoder using the output of the encoder directly as $\tilde{I}_t^{\text{ae}}$, the total loss is:
\begin{equation}
    \mathcal{L}_{total} = \mathcal{L}_{\text{pred}} + \alpha \mathcal{L}_{\text{rec}}
    = \sum_{t={L+1}}^{L+T_{pred}} \mathcal{L}(\tilde{I}_t^{\text{pred}}, I_t) + \alpha \sum_{t=1}^{L+T_{pred}} \mathcal{L}(\tilde{I}_t^{\text{ae}}, I_t) 
\end{equation}
where $\alpha$ is a hyper-parameter. We use mean-squared error loss throughout. During testing we predict an additional $T_{ext}$ frames in order to evaluate long term prediction beyond the length seen for training.

\section{Experiments}


\subsection{Physical parameter learning and future prediction}

\begin{figure}[ht]
  \centering
      \includegraphics[width=0.32\textwidth]{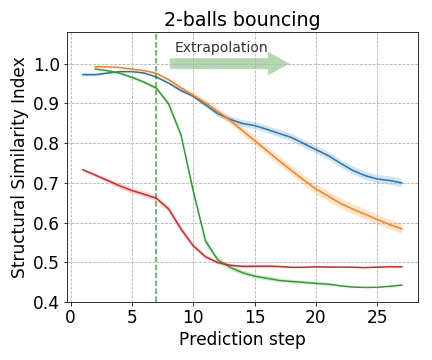}
    \includegraphics[width=0.32\textwidth]{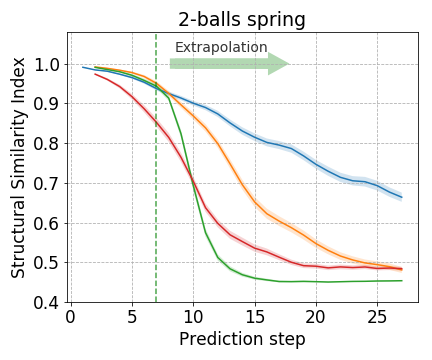}
    \includegraphics[width=0.32\textwidth]{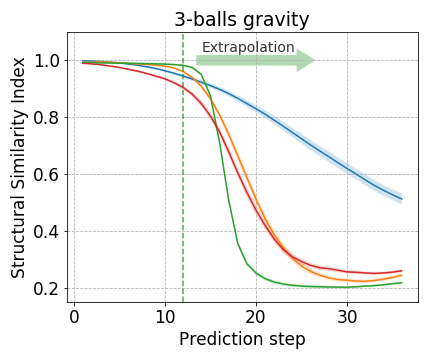}
    \includegraphics[width=0.7\textwidth]{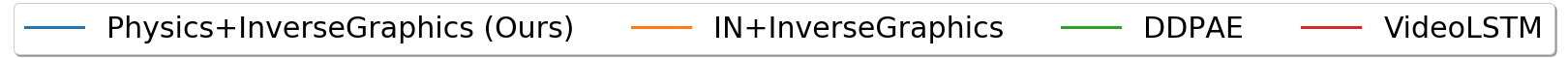}
  \caption{Frame prediction accuracy (SSI, higher is better) for the balls datasets. Left of the green dashed line corresponds to the training range, $T_{pred}$, right corresponds to extrapolation, $T_{ext}$. We outperform Interaction Networks (IN) \citep{Watters2017VisualVideo}, DDPAE \citep{Hsieh2018LearningPrediction} and VideoLSTM \citep{Srivastava2015UnsupervisedLSTMs} in  extrapolation due to incorporating explicit physics.}
  \label{fig:loss_per_step}
\end{figure}
\begin{table}[h]
\centering
\begin{tabular}{ccccc}
\hline
\textbf{Dataset}  & 
\textbf{2-balls spring} &
\textbf{2-digits spring} & 
\textbf{3-balls gravity} &
\textbf{3-balls gravity} \\ 
Parameters & $(k, \,l)$  & $(k, \,l)$ & $g$ & $m$\\ \hline 
Learned value & $(4.26,\, 6.17)$   & $(2.18,\, 12.24)$  & $65.7$ & $0.95$ \\ 
Ground-truth value & $(4.0,\, 6.0)$   & $(2.0,\, 12.0)$ & $60.0$ & $1.0$\\
\hline
\end{tabular}
\caption{Physical parameters learned from video are within 10\% of true system parameters.}
\label{table:toy_params}
\end{table}

\textbf{Setup} {} To explore learning physical parameters and evaluate long-term prediction we train our model on scenes with 5 different settings: two colored balls bouncing off the image edges; two colored balls connected by a spring; three colored balls with gravitational pull -- all on a black background; and to test greater visual complexity, we also use 2 MNSIT digits connected by a spring, on a  CIFAR background. We train using values of $(L, T_{pred}, T_{ext})$ set to $(3, 7, 20)$, $(3, 7, 20)$, $(3, 7, 20)$, $(4,12,24)$ and $(3,7,20)$, respectively. For the spring systems the physical parameters to be learned are the spring constant $k$ and equilibrium distance $l$, and for the gravitational system it is the gravity constant $g$ or mass of the objects $m$ (when learning gravity the mass if fixed, and vice-versa). In all cases we use objects with mass $m=1$. We provide the exact equations of motion used in these systems and other training details in Appendices A and B, respectively. All datasets consist of 5000 sequences for training, 500 for validation, and 500 for testing. We use a learnable ST scale parameter initialized at $s=2$ in the balls datasets and $s=1$ in the digits dataset. In these datasets we set $\theta=0$. 

\textbf{Baselines} {} We compare our model to 3 strong baselines: DDPAE \citep{Hsieh2018LearningPrediction}\footnote{Using the code provided by the authors.}, which is a generative model that uses an inverse-graphics model with black-box dynamics; VideoLSTM \citep{Srivastava2015UnsupervisedLSTMs}, which uses black-box encoding, decoding and dynamics; Interaction Network + Inverse-Graphics, which uses the same encoder and decoder as our Physics-as-Inverse-Graphics model, but where the dynamics module is an Interaction Network \citep{Battaglia2016InteractionPhysics}. The latter model allows us to compare explicit physics with relational dynamics networks, in terms of their ability to correctly capture object interactions\footnote{This baseline also serves as strong proxy for comparison with recent relational models \citep{Watters2017VisualVideo,vanSteenkiste2018RelationalInteractions}, which due to their supervision method or input-output space cannot be directly compared our model.}.

\textbf{Results} {} Table \ref{table:toy_params} shows that our model finds physical parameters close to the ground-truth values used to generate the datasets, and Figure~\ref{fig:contents} shows the contents and masks learned by the decoder. This highlights the fact that the proposed model can successfully perform unsupervised system identification from pixels. Future frame predictions for two of the systems are shown in Figure~\ref{fig:predictions}, and per-step Structural Similarity Index (SSI) 
\footnote{We choose SSI over MSE as an evaluation metric as it is more robust to pixel-level differences and alignment. \cut{ because MSE only considers differences pixel-by-pixel. If the objects are rendered very close to the true position but do not overlap, in image space this would yield already maximal MSE loss, which is undesirable.}} 
of the models on the prediction and extrapolation range are shown in Figure~ \ref{fig:loss_per_step}. While all models obtain low error in the prediction range (left of the green dashed line), our model is significantly better in  the extrapolation range. Even many steps into the future, our model's predictions are still highly accurate, unlike those of other black-box models (Figure~\ref{fig:predictions}). This shows the value of using an explicit physics model in systems  where the dynamics are non-linear yet well defined. Further rollouts are shown in Appendix C, and we encourage the reader to watch the videos for all the datasets at \url{https://sites.google.com/view/physicsasinversegraphics}.

This difference in performance is explained in part by the fact that in some of these systems the harder-to-predict parts of the dynamics do not appear during training. For example, in the gravitational system, whiplash from objects coming in close contact is seldom present in the first $K+T_{pred}$ steps given in the training set, but it happens frequently in the $T_{ext}$ extrapolation steps evaluated during testing. We do not consider this to be a failure of black-box model, but rather a consequence of the generality \textit{vs} specificity tradeoff: a model without a sufficiently strong inductive bias on the dynamics is simply not able to correctly infer close distance dynamics from long distance dynamics. \cut{Our model's physics engine, having at most a few learneable parameters, can correctly predict long term trajectories even when trained on data representing only a small fraction of all possible dynamics.}

\begin{minipage}{\textwidth}
\vspace{-0.3cm}
\hspace{0.1cm}
  \begin{minipage}[t]{0.44\textwidth}
    \begin{figure}[H]
        \centering
        \includegraphics[width=0.45\textwidth]{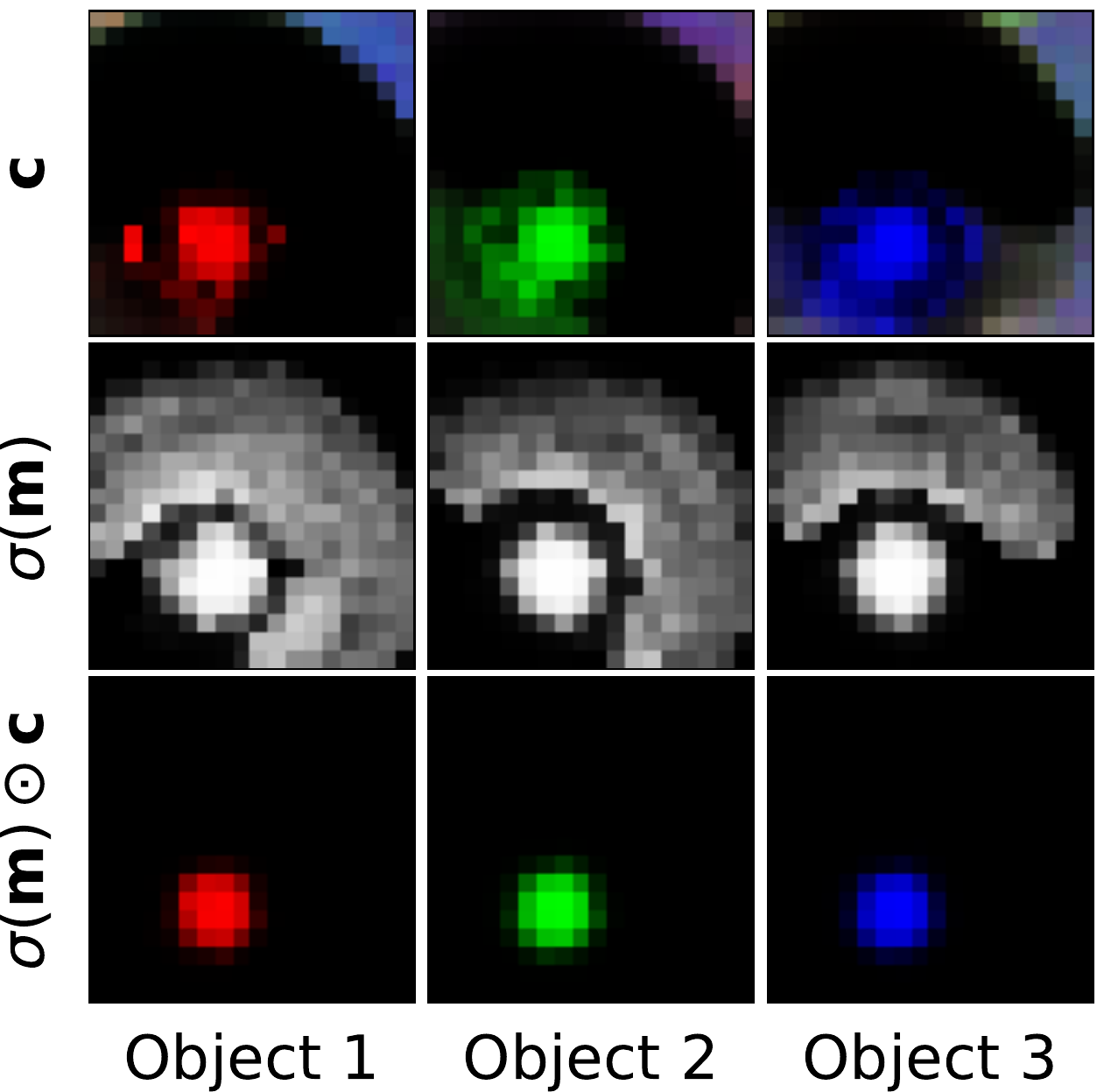}
        \hspace{0.3cm}
        \includegraphics[width=0.45\textwidth]{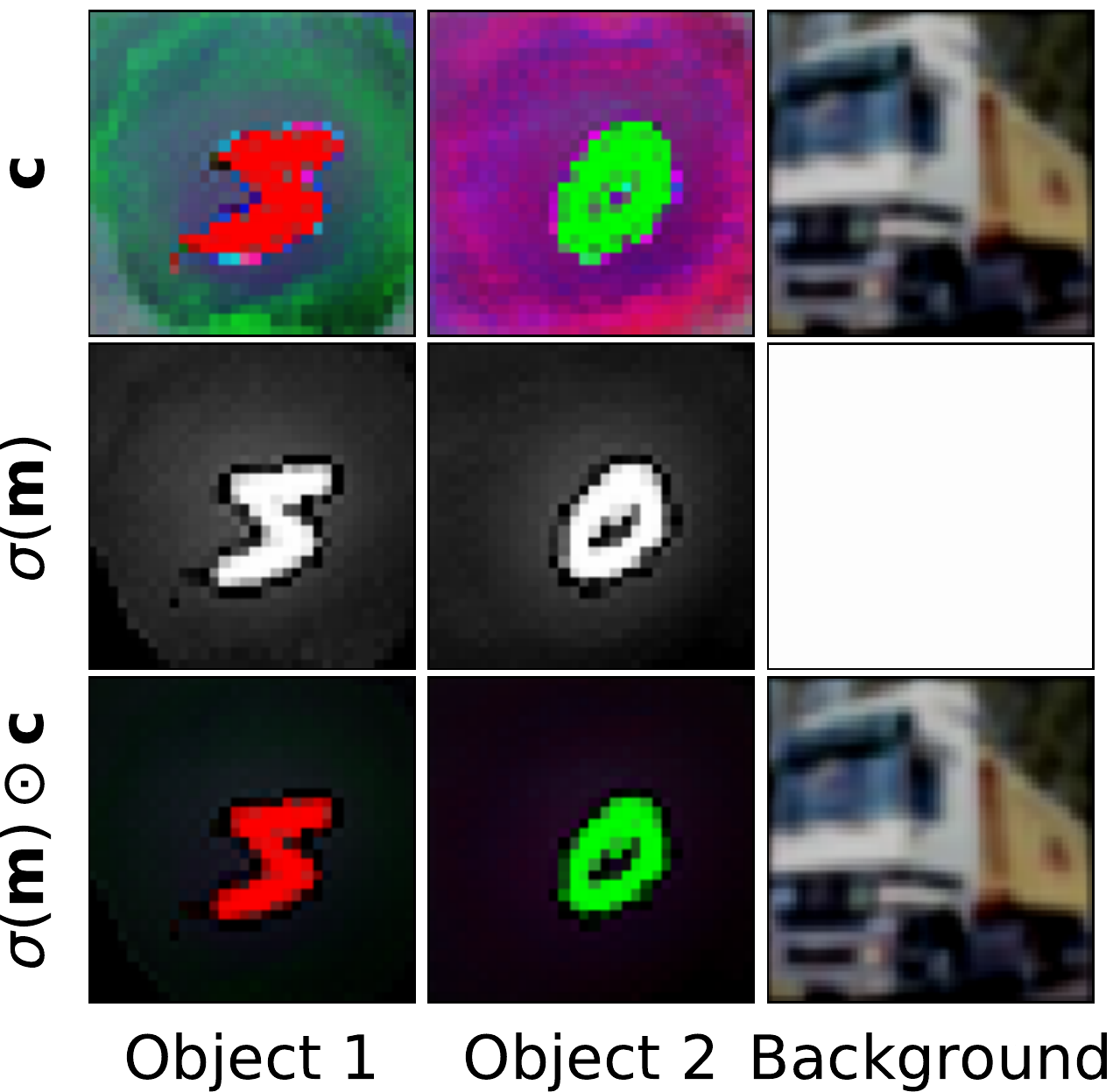}
        \vspace{-0.15cm}
        \caption{Contents and masks learned by the decoder. Object masks: $\sigma(\mbf{m})$. Objects for rendering: $\sigma(\mbf{m})\odot \mbf{c}$.  Contents and masks correctly capture each part of the scene: colored balls, MNIST digits and CIFAR background. We omit the black background learned on the balls dataset.}
      \label{fig:contents}
    \end{figure}
  \end{minipage}
  \hfill
  \begin{minipage}[t]{0.51\textwidth}
    \begin{table}[H]
        \centering
        \vspace{0.4cm}
        \begin{tabular}{ccc}
        \hline
        Train using & $\mathcal{L}_\text{pred}$ & $\mathcal{L}_\text{rec}$ \\ \hline
        only $\mathcal{L}_\text{pred}$ & 31.4 & 20.5\\ 
        separate gradients & 28.1 & 0.22 \\ 
        joint $\mathcal{L}_\text{pred}+\alpha\mathcal{L}_\text{rec}$ &  \textbf{1.39} & 0.63 \\
        black-box decoder, joint &  30.9 & 2.87 \\ \hline
        \end{tabular}
        \vspace{0.33cm}
        \caption{Test loss under different training conditions. Separate gradients: Train encoder/decoder on $\mathcal{L}_\text{rec}$, and velocity estimator and physics engine on $\mathcal{L}_\text{pred}$. Black-box decoder, joint: Joint training using a standard MLP network as the decoder. Only joint training using our coordinate-consistent decoder succeeds.} \label{table:ablation}
    \end{table}
  \end{minipage}
  \hspace{0.1cm}
\end{minipage}

\textbf{Ablation studies} {}
\label{sec:ablation}
Since the encoder and decoder must discover the objects present in the image and the corresponding locations, one might assume that the velocity estimator and physics engine  could be learned using only the prediction loss, and encoder/decoder using only the static autoencoder loss, i.e., without joint training. In Table~\ref{table:ablation} we compare the performance of four variants on the 3-ball gravity dataset: joint training using only the prediction loss; joint training using the prediction and autoencoder losses; training the encoder/decoder on the autoencoder loss and the velocity estimator and physics engine on the prediction loss; and joint training, but using an MLP black-box decoder.


We can see that only joint prediction and autoencoder loss obtain satisfactory performance, and that the use of the proposed coordinate-consistent decoder is critical. 
The prediction loss is essential in order for the model to learn encoders/decoders whose content and masks can be correctly used by the physics engine. This can be understood by considering how object interaction influences the decoder.
In the gravitational system, the forces between objects depend on their distances, so if the objects swap locations, the forces must be the same. If the content/mask learned for each object are centered differently relative to its template center, rendering the objects at positions $[x,y]$ and $[w,z]$, or $[w,z]$ and $[x,y]$ will produce different distances between these two objects in image space. This violates the permutation invariance property of the system.  Learning the encoder/decoder along with the velocity estimator and physics engine on the prediction loss allows the encoder and decoder to learn locations and contents/masks that satisfy the characteristics of the system and allows the physics to be learned correctly. In Appendix D we perform further ablations on the decoder architecture and its ability to correctly render objects in regions of the image not seen during training.

\subsection{Vision-based model-predictive control (MPC)}



\textbf{Tasks} {} One of the main applications of our method is to identify the (actuated) dynamical parameters and states of a physical system from video, which enables vision-based planning and control. Here we apply it to the pendulum from OpenAI Gym \citep{Brockman2016OpenAIGym} -- one typically solved from proprioceptive state, not pixels. For training we collect 5000 sequences of 14 frames with random initialization ($\Dot{\theta}_0 \sim \text{Unif}(-6, 6)$) and actions ($u_t \sim \text{Unif}(-2, 2)$). The physical parameters to learn are gravity $g=10.0$ and actuation coefficient $a=1.0$. We use $K=4$ and $T_{pred}=10$.
We use the trained MPC model  as follows. At every step, the previous 4 frames are passed to the encoder and velocity nets to estimate $[\theta_t, \Dot{\theta}_t]$. This is passed  to the physics engine with  learned parameters $g$ and $a$. We perform 100-step model-predictive control using the cross entropy method \citep{Rubinstein1997OptimizationEvents}, exactly as described in \cite{hafner2018Planet}, setting vertical position and zero velocity as the goal. 
\textbf{Baselines} {} We compare our model to an oracle model, which has the true physical parameters and access to the true pendulum position and velocity (not vision-based), as well as a concurrent state-of-the art model-based RL method (PlaNet \citep{hafner2018Planet}), and a model-free\footnote{DDPG, TRPO and PPO learned from pixels failed to solve the pendulum, highlighting the complexity of the vision-based pendulum control task and brittleness of model-free reinforcement learning strategies.} deep deterministic policy gradient (DDPG) agent \citep{lillicrap2015continuous}. To provide an equivalent comparison to our model, we train PlaNet on random episodes.

\begin{figure}[tb]
  \centering
      \includegraphics[width=0.35\textwidth]{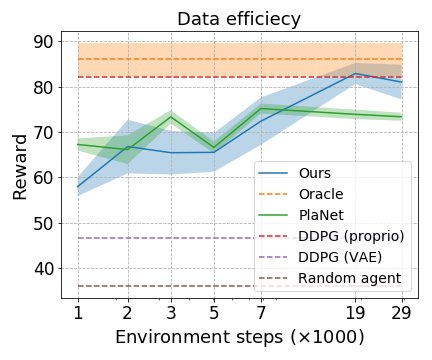}
    \includegraphics[width=0.35\textwidth]{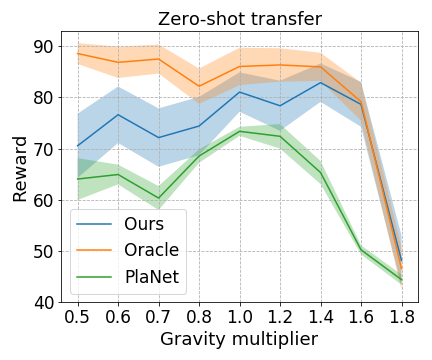}
    \includegraphics[width=0.285\textwidth]{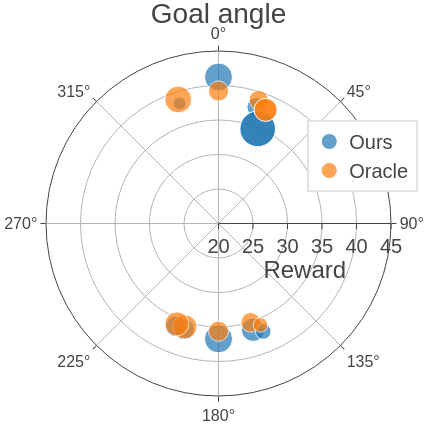}
    
    \vspace{0.2cm}
    
    \includegraphics[width=0.99\textwidth]{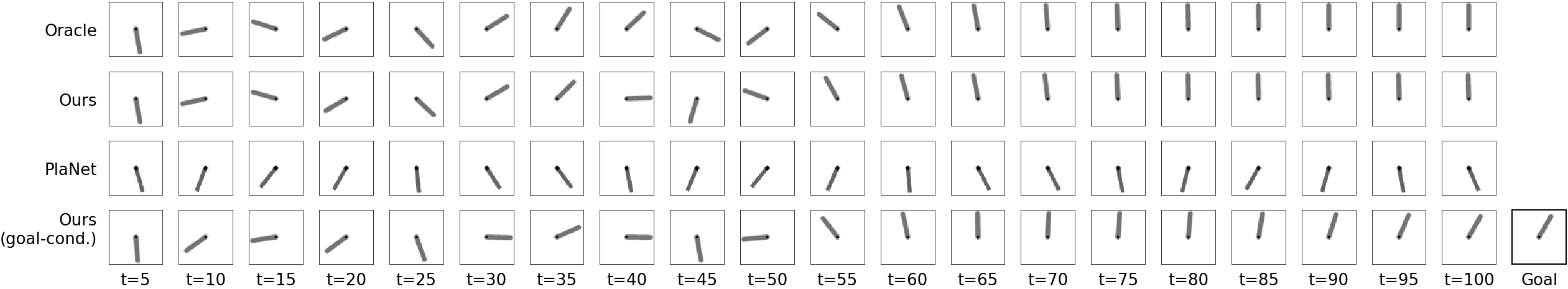}
  \caption{\textbf{Top:} Comparison between our model and PlaNet \cite{hafner2018Planet} in terms of learning sample efficiency \textbf{(left)}. Explicit physics allows reasoning for zero-shot adaptation to domain-shift in gravity (\textbf{center}) and goal-driven control to balance the pendulum in any position (\textbf{right}). DDPG (VAE) corresponds to a DDPG agent trained on the latent space of an autoencoder (trained with 320k images) after 80k steps. DDPG (proprio) corresponds to an agent trained from proprioception after 30k steps. \textbf{Bottom:} The first 3 rows show a zero-shot counterfactual episode with a gravity multiplier of 1.4 for an oracle, our model and planet, with vertical as the target position (as trained). The last row shows an episode using a goal image to infer the non-vertical goal state.}
  \label{fig:pendulum_results}
\end{figure}

\textbf{Results} {} In terms of system identification, our model recovers the correct gravity $(g=9.95)$ and force coefficient $(a=0.99)$ values from vision alone, which is a prerequisite to perform correct planning and control.
Figure~\ref{fig:pendulum_results} (top-left) highlights the data efficiency of our method, which is comparable to PlaNet, while being dramatically faster than DDPG from pixels. Importantly, the interpretability of the explicit physics in our model provides some unique capabilities. We can perform simple counter-factual physical reasoning such as `\emph{How should I adapt my control policy if gravity was increased?}', which enables zero-shot adaptation to new environmental parameters. Figure~\ref{fig:pendulum_results} (top-middle) shows that our model can exploit such reasoning to succeed immediately over a wide range of gravities with no re-training. Similarly, while the typical inverted pendulum goal is to balance the pendulum upright, interpretable physics means that this is only one point in a space of potential goals.  Figure~\ref{fig:pendulum_results} (top-right) evaluates the goal-paramaterized control enabled by our model. Any feasible target angle specified can be directly reached by the controller. There is extrapolative generalisation across the space of goals even though only one goal (vertical) was seen during training. Importantly, these last two capabilities are provided automatically by our model due to its disentangled interpretable representation, but cannot be achieved without further adaptive learning by alternatives that are reward-based \citep{mnih2016asynchronous} or rely on implicit physics \citep{hafner2018Planet}.

\section{Limitations}

Alhough the approach presented here shows promising results in terms of physical parameter estimation, long-term video prediction and MPC, a number of limitations need to be overcome for real-world application.

\textbf{Templates as object representation} {} Though the assumption that every scene in a dataset is a combination of learnable templates is a common one in the literature (c.f. \cite{Tieleman} for an extensive study on this), this is insufficient to model real-world scenes. For example, applying physics-as-inverse-graphics to the Physics101 dataset would require representing objects using a latent appearance representation that could be used by the decoder \citep{Eslami}. This would introduce new modelling challenges, requiring object tracking to keep correct object identity associations \citep{Kosiorek2018SequentialObjects}. In this work we simplify this problem by assuming that objects are visually distinct throughout the dataset, though this does not detract from the essential contributions of the paper.

\textbf{Rigid sequence to sequence architecture} {} In this work we used a sequence-to-sequence architecture, with a fixed number of input steps. This architectural choice (inspired by \cite{Watters2017VisualVideo}), prevents the model from updating its state beliefs if given additional input frames later in the sequence. Formulating the current model in a probabilistic manner that would allow for state/parameter filtering and smoothing at inference time is a promising direction of future work.

\textbf{Static background assumption} {} Many scenes of interest do not follow the assumption that the only moving objects in the scene are the objects of interest (even though this assumption is widely used). Adapting our model to varying scene backgrounds would require additional components to discern which parts of the scene follow the dynamics assumed by the physics engine, in order to correctly perform object discovery. This is a challenging problem, but we believe it would greatly increase the range of applications of the ideas presented here.

\section{Conclusion}
Physics-as-inverse graphics provides a valuable mechanism to integrate inductive bias about physical data generating processes into learning. This allows unsupervised object tracking and system identification, in addition to sample efficient, generalisable and flexible control. However, incorporating this structure into lightly supervised deep learning models has proven challenging to date. We introduced a model that accomplishes this, relying on a  coordinate-consistent decoder that enables image reconstruction from physics. We have shown that our model is able to perform accurate long term prediction and that it can be used to learn the dynamics of an actuated system, allowing us to perform vision-based model-predictive control.


\bibliography{references,references_2}

\begin{thebibliography}{3}
\providecommand{\natexlab}[1]{#1}
\providecommand{\url}[1]{\texttt{#1}}
\expandafter\ifx\csname urlstyle\endcsname\relax
  \providecommand{\doi}[1]{doi: #1}\else
  \providecommand{\doi}{doi: \begingroup \urlstyle{rm}\Url}\fi

\bibitem[Bengio \& LeCun(2007)Bengio and LeCun]{Bengio+chapter2007}
Yoshua Bengio and Yann LeCun.
\newblock Scaling learning algorithms towards {AI}.
\newblock In \emph{Large Scale Kernel Machines}. MIT Press, 2007.

\bibitem[Goodfellow et~al.(2016)Goodfellow, Bengio, Courville, and
  Bengio]{goodfellow2016deep}
Ian Goodfellow, Yoshua Bengio, Aaron Courville, and Yoshua Bengio.
\newblock \emph{Deep learning}, volume~1.
\newblock MIT Press, 2016.

\bibitem[Hinton et~al.(2006)Hinton, Osindero, and Teh]{Hinton06}
Geoffrey~E. Hinton, Simon Osindero, and Yee~Whye Teh.
\newblock A fast learning algorithm for deep belief nets.
\newblock \emph{Neural Computation}, 18:\penalty0 1527--1554, 2006.

\end{thebibliography}


\begin{thebibliography}{53}
\providecommand{\natexlab}[1]{#1}
\providecommand{\url}[1]{\texttt{#1}}
\expandafter\ifx\csname urlstyle\endcsname\relax
  \providecommand{\doi}[1]{doi: #1}\else
  \providecommand{\doi}{doi: \begingroup \urlstyle{rm}\Url}\fi

\bibitem[Barth-Maron et~al.(2018)Barth-Maron, Hoffman, Budden, Dabney, Horgan,
  Muldal, Heess, and Lillicrap]{barth2018distributed}
Gabriel Barth-Maron, Matthew~W Hoffman, David Budden, Will Dabney, Dan Horgan,
  Alistair Muldal, Nicolas Heess, and Timothy Lillicrap.
\newblock Distributed distributional deterministic policy gradients.
\newblock \emph{ICLR}, 2018.

\bibitem[Battaglia et~al.(2016)Battaglia, Pascanu, Lai, Rezende, and
  Kavukcuoglu]{Battaglia2016InteractionPhysics}
Peter~W. Battaglia, Razvan Pascanu, Matthew Lai, Danilo Rezende, and Koray
  Kavukcuoglu.
\newblock {Interaction Networks for Learning about Objects, Relations and
  Physics}.
\newblock In \emph{NIPS}, 2016.

\bibitem[Belbute-Peres et~al.(2018)Belbute-Peres, Smith, Allen, Tenenbaum, and
  Kolter]{Belbute-Peres2018End-to-EndControl}
Filipe De~A Belbute-Peres, Kevin~A Smith, Kelsey~R Allen, Joshua~B Tenenbaum,
  and J~Zico Kolter.
\newblock {End-to-End Differentiable Physics for Learning and Control}.
\newblock In \emph{NIPS}, 2018.

\bibitem[Brockman et~al.(2016)Brockman, Cheung, Pettersson, Schneider,
  Schulman, Tang, and Zaremba]{Brockman2016OpenAIGym}
Greg Brockman, Vicki Cheung, Ludwig Pettersson, Jonas Schneider, John Schulman,
  Jie Tang, and Wojciech Zaremba.
\newblock {OpenAI Gym}.
\newblock 2016.

\bibitem[Burgess et~al.(2019)Burgess, Matthey, Watters, Kabra, Higgins,
  Botvinick, and Lerchner]{Burgess2019MONet:Representation}
Christopher~P Burgess, Loic Matthey, Nicholas Watters, Rishabh Kabra, Irina
  Higgins, Matt Botvinick, and Alexander Lerchner.
\newblock {MONet: Unsupervised Scene Decomposition and Representation}.
\newblock \emph{CoRR}, abs/1901.11390, 2019.

\bibitem[Burke et~al.(2019)Burke, Penkov, and Ramamoorthy]{Burke19Explanation}
Michael Burke, Svetlin Penkov, and Subramanian Ramamoorthy.
\newblock From explanation to synthesis: Compositional program induction for
  learning from demonstration.
\newblock \emph{Robotics: Science and Systems (R:SS)}, 2019.

\bibitem[Byravan et~al.(2018)Byravan, Leeb, Meier, and
  Fox]{Byravan2018SE3-Pose-Nets:Control}
Arunkumar Byravan, Felix Leeb, Franziska Meier, and Dieter Fox.
\newblock {SE3-Pose-Nets: Structured Deep Dynamics Models for Visuomotor
  Planning and Control}.
\newblock In \emph{ICRA}, 2018.

\bibitem[Chang et~al.(2017)Chang, Ullman, Torralba, and
  Tenenbaum]{Chang2017ADynamics}
Michael~B Chang, Tomer Ullman, Antonio Torralba, and Joshua~B Tenenbaum.
\newblock {A Compositional Object-Based Approach to Learning Physical
  Dynamics}.
\newblock In \emph{ICLR}, 2017.

\bibitem[Chen et~al.(2018)Chen, Rubanova, Bettencourt, and
  Duvenaud]{Chen2018NeuralEquations}
Ricky T~Q Chen, Yulia Rubanova, Jesse Bettencourt, and David Duvenaud.
\newblock {Neural Ordinary Differential Equations}.
\newblock In \emph{NIPS}, 2018.

\bibitem[Degrave et~al.(2016)Degrave, Hermans, Dambre, and
  wyffels]{Degrave2016ARobotics}
Jonas Degrave, Michiel Hermans, Joni Dambre, and Francis wyffels.
\newblock {A Differentiable Physics Engine for Deep Learning in Robotics}.
\newblock 11 2016.

\bibitem[Deisenroth \& Rasmussen(2011)Deisenroth and
  Rasmussen]{deisenroth2011pilco}
Marc Deisenroth and Carl~E Rasmussen.
\newblock Pilco: A model-based and data-efficient approach to policy search.
\newblock In \emph{ICML}, 2011.

\bibitem[Ehrhardt et~al.(2018)Ehrhardt, Monszpart, Vedaldi, and
  Mitra]{Ehrhardt2018UnsupervisedObservations}
Sebastien Ehrhardt, Aron Monszpart, Andrea Vedaldi, and Niloy Mitra.
\newblock {Unsupervised Intuitive Physics from Visual Observations}.
\newblock \emph{CoRR}, abs/1805.08095, 2018.

\bibitem[Ellis et~al.(2018)Ellis, Ritchie, Solar-Lezama, and
  Tenenbaum]{Ellis2018LearningImages}
Kevin Ellis, Daniel Ritchie, Armando Solar-Lezama, and Joshua~B. Tenenbaum.
\newblock {Learning to Infer Graphics Programs from Hand-Drawn Images}.
\newblock In \emph{NIPS}, 2018.

\bibitem[Eslami et~al.(2016)Eslami, Heess, Weber, Tassa, Kavukcuoglu, and
  Hinton]{Eslami}
S~M~Ali Eslami, Nicolas Heess, Theophane Weber, Yuval Tassa, Koray Kavukcuoglu,
  and Geoffrey~E Hinton.
\newblock {Attend, Infer, Repeat: Fast Scene Understanding with Generative
  Models}.
\newblock In \emph{NIPS}, 2016.

\bibitem[Finn et~al.(2016)Finn, Goodfellow, and
  Levine]{Finn2016UnsupervisedPrediction}
Chelsea Finn, Ian Goodfellow, and Sergey Levine.
\newblock {Unsupervised Learning for Physical Interaction through Video
  Prediction}.
\newblock In \emph{NIPS}, 2016.

\bibitem[Fraccaro et~al.(2017)Fraccaro, Kamronn, Paquet, and
  Winther]{Fraccaro2017ALearning}
Marco Fraccaro, Simon Kamronn, Ulrich Paquet, and Ole Winther.
\newblock {A Disentangled Recognition and Nonlinear Dynamics Model for
  Unsupervised Learning}.
\newblock In \emph{NIPS}, 2017.

\bibitem[Fragkiadaki et~al.(2016)Fragkiadaki, Agrawal, Levine, and
  Malik]{Fragkiadaki2016LearningBilliards}
Katerina Fragkiadaki, Pulkit Agrawal, Sergey Levine, and Jitendra Malik.
\newblock {Learning Visual Predictive Models of Physics for Playing Billiards}.
\newblock In \emph{ICLR}, 2016.

\bibitem[Fukushima(1980)]{fukushima1980neocognitron}
Kunihiko Fukushima.
\newblock Neocognitron: A self-organizing neural network model for a mechanism
  of pattern recognition unaffected by shift in position.
\newblock \emph{Biological cybernetics}, 36\penalty0 (4):\penalty0 193--202,
  1980.

\bibitem[Greff et~al.(2019)Greff, Kaufman, Kabra, Watters, Burgess, Zoran,
  Matthey, Botvinick, and Lerchner]{Greff2019Multi-ObjectInference}
Klaus Greff, Raphaël~Lopez Kaufman, Rishabh Kabra, Nick Watters, Chris
  Burgess, Daniel Zoran, Loic Matthey, Matthew Botvinick, and Alexander
  Lerchner.
\newblock {Multi-Object Representation Learning with Iterative Variational
  Inference}.
\newblock In \emph{ICML}, 2019.

\bibitem[Hafner et~al.(2019)Hafner, Lillicrap, Fischer, Villegas, Ha, Lee, and
  Davidson]{hafner2018Planet}
Danijar Hafner, Timothy Lillicrap, Ian Fischer, Ruben Villegas, David Ha,
  Honglak Lee, and James Davidson.
\newblock Learning latent dynamics for planning from pixels.
\newblock \emph{ICML}, 2019.

\bibitem[Hinton et~al.(2012)Hinton, Srivastava, and
  Swersky]{Hinton2012LectureDescent}
Geoffrey Hinton, Nitish Srivastava, and Kevin Swersky.
\newblock {Lecture 6a: Overview of mini­-batch gradient descent}.
\newblock In \emph{Neural Networks for Machine Learning}, 2012.

\bibitem[Hinton et~al.(2011)Hinton, Krizhevsky, and Wang]{Hinton}
Geoffrey~E. Hinton, Alex Krizhevsky, and Sida~D. Wang.
\newblock {Transforming auto-encoders}.
\newblock In \emph{ICANN}, pp.\  44--51, 2011.

\bibitem[Hochreiter \& Schmidhuber(1997)Hochreiter and
  Schmidhuber]{hochreiter1997long}
Sepp Hochreiter and J{\"u}rgen Schmidhuber.
\newblock Long short-term memory.
\newblock \emph{Neural computation}, 9\penalty0 (8):\penalty0 1735--1780, 1997.

\bibitem[Hsieh et~al.(2018)Hsieh, Liu, Huang, Fei-Fei, and
  Niebles]{Hsieh2018LearningPrediction}
Jun-Ting Hsieh, Bingbin Liu, De-An Huang, Li~Fei-Fei, and Juan~Carlos Niebles.
\newblock {Learning to Decompose and Disentangle Representations for Video
  Prediction}.
\newblock In \emph{NIPS}, 2018.

\bibitem[Huang \& Murphy(2016)Huang and Murphy]{Huang}
Jonathan Huang and Kevin Murphy.
\newblock {Efficient Inference in Occlusion-Aware Generative Models of Images}.
\newblock In \emph{ICLR Workshop}, 2016.

\bibitem[Jaderberg et~al.(2015)Jaderberg, Simonyan, Zisserman, and
  Kavukcuoglu]{Jaderberg}
Max Jaderberg, Karen Simonyan, Andrew Zisserman, and Koray Kavukcuoglu.
\newblock {Spatial Transformer Networks}.
\newblock In \emph{NIPS}, 2015.

\bibitem[Janner et~al.(2019)Janner, Levine, Freeman, Tenenbaum, Finn, and
  Wu]{Janner2019ReasoningPlanning}
Michael Janner, Sergey Levine, William~T Freeman, Joshua~B Tenenbaum, Chelsea
  Finn, and Jiajun Wu.
\newblock {Reasoning About Physical Interactions with Object-Oriented
  Prediction and Planning}.
\newblock In \emph{ICLR}, 2019.

\bibitem[Jonschkowski et~al.(2017)Jonschkowski, Hafner, Scholz, and
  Riedmiller]{Jonschkowski2017PVEs:Representations}
Rico Jonschkowski, Roland Hafner, Jonathan Scholz, and Martin Riedmiller.
\newblock {PVEs: Position-Velocity Encoders for Unsupervised Learning of
  Structured State Representations}.
\newblock \emph{CoRR}, abs/1705.09805, 2017.

\bibitem[Kosiorek et~al.(2018)Kosiorek, Kim, Posner, and
  Whye~Teh]{Kosiorek2018SequentialObjects}
Adam~R Kosiorek, Hyunjik Kim, Ingmar Posner, and Yee Whye~Teh.
\newblock {Sequential Attend, Infer, Repeat: Generative Modelling of Moving
  Objects}.
\newblock In \emph{NIPS}, 2018.

\bibitem[Kulkarni et~al.(2015)Kulkarni, Whitney, Kohli, and
  Tenenbaum]{Kulkarni}
Tejas~D Kulkarni, Will Whitney, Pushmeet Kohli, and Joshua~B Tenenbaum.
\newblock {Deep Convolutional Inverse Graphics Network}.
\newblock In \emph{NIPS}, 2015.

\bibitem[Lillicrap et~al.(2016)Lillicrap, Hunt, Pritzel, Heess, Erez, Tassa,
  Silver, and Wierstra]{lillicrap2015continuous}
Timothy~P Lillicrap, Jonathan~J Hunt, Alexander Pritzel, Nicolas Heess, Tom
  Erez, Yuval Tassa, David Silver, and Daan Wierstra.
\newblock Continuous control with deep reinforcement learning.
\newblock \emph{ICLR}, 2016.

\bibitem[Mania et~al.(2018)Mania, Guy, and Recht]{ManiaRech18}
Horia Mania, Aurelia Guy, and Benjamin Recht.
\newblock Simple random search provides a competitive approach to reinforcement
  learning.
\newblock \emph{NIPS}, 2018.

\bibitem[Mnih et~al.(2016)Mnih, Badia, Mirza, Graves, Lillicrap, Harley,
  Silver, and Kavukcuoglu]{mnih2016asynchronous}
Volodymyr Mnih, Adria~Puigdomenech Badia, Mehdi Mirza, Alex Graves, Timothy
  Lillicrap, Tim Harley, David Silver, and Koray Kavukcuoglu.
\newblock Asynchronous methods for deep reinforcement learning.
\newblock In \emph{ICML}, 2016.

\bibitem[Penkov \& Ramamoorthy(2019)Penkov and Ramamoorthy]{penkov2019learning}
Svetlin Penkov and Subramanian Ramamoorthy.
\newblock Learning programmatically structured representations with perceptor
  gradients.
\newblock \emph{ICLR}, 2019.

\bibitem[Rezende et~al.(2016)Rezende, Mohamed, Danihelka, Gregor, and
  Wierstra]{Rezende2016One-ShotModels}
Danilo~J. Rezende, Shakir Mohamed, Ivo Danihelka, Karol Gregor, and Daan
  Wierstra.
\newblock {One-Shot Generalization in Deep Generative Models}.
\newblock In \emph{ICML}, 2016.

\bibitem[Romaszko et~al.(2017)Romaszko, Williams, Moreno, and
  Kohli]{Romaszko2017Vision-as-Inverse-Graphics:Image}
Lukasz Romaszko, Christopher K~I Williams, Pol Moreno, and Pushmeet Kohli.
\newblock {Vision-as-Inverse-Graphics: Obtaining a Rich 3D Explanation of a
  Scene from a Single Image}.
\newblock In \emph{ICCV}, 2017.

\bibitem[Ronneberger et~al.(2015)Ronneberger, Fischer, and
  Brox]{Ronneberger2015U-Net:Segmentation}
Olaf Ronneberger, Philipp Fischer, and Thomas Brox.
\newblock {U-Net: Convolutional Networks for Biomedical Image Segmentation}.
\newblock In \emph{MICCAI}, 2015.

\bibitem[Rubinstein(1997)]{Rubinstein1997OptimizationEvents}
Reuven~Y. Rubinstein.
\newblock {Optimization of computer simulation models with rare events}.
\newblock \emph{EJOR}, 1997.

\bibitem[Srivastava et~al.(2015)Srivastava, Mansimov, and
  Salakhutdinov]{Srivastava2015UnsupervisedLSTMs}
Nitish Srivastava, Elman Mansimov, and Ruslan Salakhutdinov.
\newblock {Unsupervised Learning of Video Representations using LSTMs}.
\newblock In \emph{ICML}, 2015.

\bibitem[Stewart \& Ermon(2017)Stewart and
  Ermon]{Stewart2017Label-FreeKnowledge}
Russell Stewart and Stefano Ermon.
\newblock {Label-Free Supervision of Neural Networks with Physics and Domain
  Knowledge}.
\newblock In \emph{AAAI}, 2017.

\bibitem[Tieleman(2014)]{Tieleman}
Tijmen Tieleman.
\newblock {Optimizing Neural Networks that Generate Images}.
\newblock \emph{PhD thesis}, 2014.

\bibitem[van Steenkiste et~al.(2018)van Steenkiste, Chang, Greff, and
  Schmidhuber]{vanSteenkiste2018RelationalInteractions}
Sjoerd van Steenkiste, Michael Chang, Klaus Greff, and Jürgen Schmidhuber.
\newblock {Relational Neural Expectation Maximization: Unsupervised Discovery
  of Objects and their Interactions}.
\newblock In \emph{ICLR}, 2018.

\bibitem[Watters et~al.(2017)Watters, Tacchetti, Weber, Pascanu, Battaglia, and
  Zoran]{Watters2017VisualVideo}
Nicholas Watters, Andrea Tacchetti, Théophane Weber, Razvan Pascanu, Peter
  Battaglia, and Daniel Zoran.
\newblock {Visual Interaction Networks: Learning a Physics Simulator from
  Video}.
\newblock In \emph{NIPS}, 2017.

\bibitem[Watters et~al.(2019{\natexlab{a}})Watters, Matthey, Bosnjak, Burgess,
  and Lerchner]{Watters2019COBRA:Exploration}
Nicholas Watters, Loic Matthey, Matko Bosnjak, Christopher~P. Burgess, and
  Alexander Lerchner.
\newblock {COBRA: Data-Efficient Model-Based RL through Unsupervised Object
  Discovery and Curiosity-Driven Exploration}.
\newblock \emph{CoRR}, abs/1905.09275, 2019{\natexlab{a}}.

\bibitem[Watters et~al.(2019{\natexlab{b}})Watters, Matthey, Burgess, and
  Lerchner]{Watters2019SpatialVAEs}
Nicholas Watters, Loic Matthey, Christopher~P. Burgess, and Alexander Lerchner.
\newblock {Spatial Broadcast Decoder: A Simple Architecture for Learning
  Disentangled Representations in VAEs}.
\newblock 2019{\natexlab{b}}.

\bibitem[Williams(1992)]{Williams1992}
Ronald~J. Williams.
\newblock {Simple Statistical Gradient-Following Algorithms for Connectionist
  Reinforcement Learning}.
\newblock \emph{Machine Learning}, 8:\penalty0 229--256, 1992.

\bibitem[Wu et~al.(2015)Wu, Yildirim, Lim, Freeman, and
  Tenenbaum]{Wu2015GalileoLearning}
Jiajun Wu, Ilker Yildirim, J.J. Lim, W.T. Freeman, and J.B. Tenenbaum.
\newblock {Galileo : Perceiving Physical Object Properties by Integrating a
  Physics Engine with Deep Learning}.
\newblock In \emph{NIPS}, 2015.

\bibitem[Wu et~al.(2016)Wu, Lim, Zhang, Tenenbaum, and
  Freeman]{Wu2016PhysicsVideos}
Jiajun Wu, Joseph~J Lim, Hongyi Zhang, Joshua~B Tenenbaum, and William~T
  Freeman.
\newblock {Physics 101: Learning physical object properties from unlabeled
  videos}.
\newblock In \emph{BMVC}, 2016.

\bibitem[Wu et~al.(2017{\natexlab{a}})Wu, Csail, Tenenbaum, and
  Kohli]{Wu2017NeuralDe-rendering}
Jiajun Wu, Mit Csail, Joshua~B Tenenbaum, and Pushmeet Kohli.
\newblock {Neural Scene De-rendering}.
\newblock In \emph{CVPR}, 2017{\natexlab{a}}.

\bibitem[Wu et~al.(2017{\natexlab{b}})Wu, Lu, Kohli, Freeman, and
  Tenenbaum]{Wu2017LearningDe-animation}
Jiajun Wu, Erika Lu, Pushmeet Kohli, William~T Freeman, and Joshua~B Tenenbaum.
\newblock {Learning to See Physics via Visual De-animation}.
\newblock In \emph{NIPS}, 2017{\natexlab{b}}.

\bibitem[Xu et~al.(2019)Xu, Liu, Sun, Research, Murphy, Freeman, Tenenbaum, and
  Wu]{Xu2019UnsupervisedDynamics}
Zhenjia Xu, Zhijian Liu, Chen Sun, Google Research, Kevin Murphy, William~T
  Freeman, Joshua~B Tenenbaum, and Jiajun Wu.
\newblock {Unsupervised Discovery of Parts, Structure, and Dynamics}.
\newblock In \emph{ICLR}, 2019.

\bibitem[Zheng et~al.(2018)Zheng, Luo, Wu, and
  Tenenbaum]{Zheng2018UnsupervisedNetworks}
David Zheng, Vinson Luo, Jiajun Wu, and Joshua~B Tenenbaum.
\newblock {Unsupervised Learning of Latent Physical Properties Using
  Perception-Prediction Networks}.
\newblock In \emph{UAI}, 2018.

\bibitem[Zhu et~al.(2018)Zhu, Huang, and
  Zhang]{Zhu2018Object-OrientedPredictor}
Guangxiang Zhu, Zhiao Huang, and Chongjie Zhang.
\newblock {Object-Oriented Dynamics Predictor}.
\newblock In \emph{NIPS}, 2018.

\end{thebibliography}
\bibliographystyle{iclr2020_conference}

\appendix
\section{System descriptions}
\label{physics}

In this section we describe the equations of motion used for each system.

\textbf{2-balls and 2-digits spring} {} 
The force applied on object $i$ by object $j$ follows Hooke's law:
\begin{equation}
    \vec{F}_{i,j} = -k\, (\vec{p}_i-\vec{p}_j) - l \frac{\vec{p}_i-\vec{p}_j}{|\vec{p}_i-\vec{p}_j|}. 
\end{equation}
Each step corresponds to an interval $\Delta t=0.3$.

\textbf{3-balls gravity} {}
The force applied on object $i$ by object $j$ follows Newton's law of gravity:
\begin{equation}
    \vec{F}_{i,j} = -g \,m_i m_j \frac{\vec{p}_i-\vec{p}_j}{|\vec{p}_i-\vec{p}_j|^3} 
\end{equation}
where the masses are set to 1. Each step corresponds to an interval $\Delta t=0.5$.

\textbf{Pendulum} {} 
The pendulum follows the equations used by the OpenAI Gym environment:
\begin{equation}
    \vec{F} = -\frac{3}{2}g \sin(\theta + \pi) + 3u
\end{equation}
where $u$ is the action. Each step corresponds to an interval $\Delta t=0.05$. In the physics engine used by the model we introduce an extra actuation coefficient $a$ to be learned along with $g$:
\begin{equation}
    \vec{F} = -\frac{3}{2}g \sin(\theta + \pi) + a\cdot u
\end{equation}

\section{Training details and hyperparameters}

For all datasets we use RMSProp \cite{Hinton2012LectureDescent} with an initial learning rate of $3\times 10^{-4}$. For the balls and digits datasets we train for 500 epochs with $\alpha=2$, and divide the learning rate by 5 after 375 epochs. For the pendulum data we train for 1000 epochs using $\alpha=3$, but divide the learning rate by 5 after 500 epochs. The image sizes are $32\times32$ for the 2-balls bouncing and spring,  $36\times36$ for the 3-balls gravity, $64\times64$ for the 2-digits spring, and $64\times64$ grayscale for the pendulum.

The content and mask variables are the output of a neural network with a constant array of 1s as input and 1 hidden layer with 200 units and tanh activation. We found this easier to train rather than having the contents and masks as trainable variables themselves. 

\newpage

\section{Additional rollouts for each dataset}

\subsubsection*{3-balls gravity}
\begin{figure}[H]
  \centering
    \includegraphics[width=0.80\textwidth]{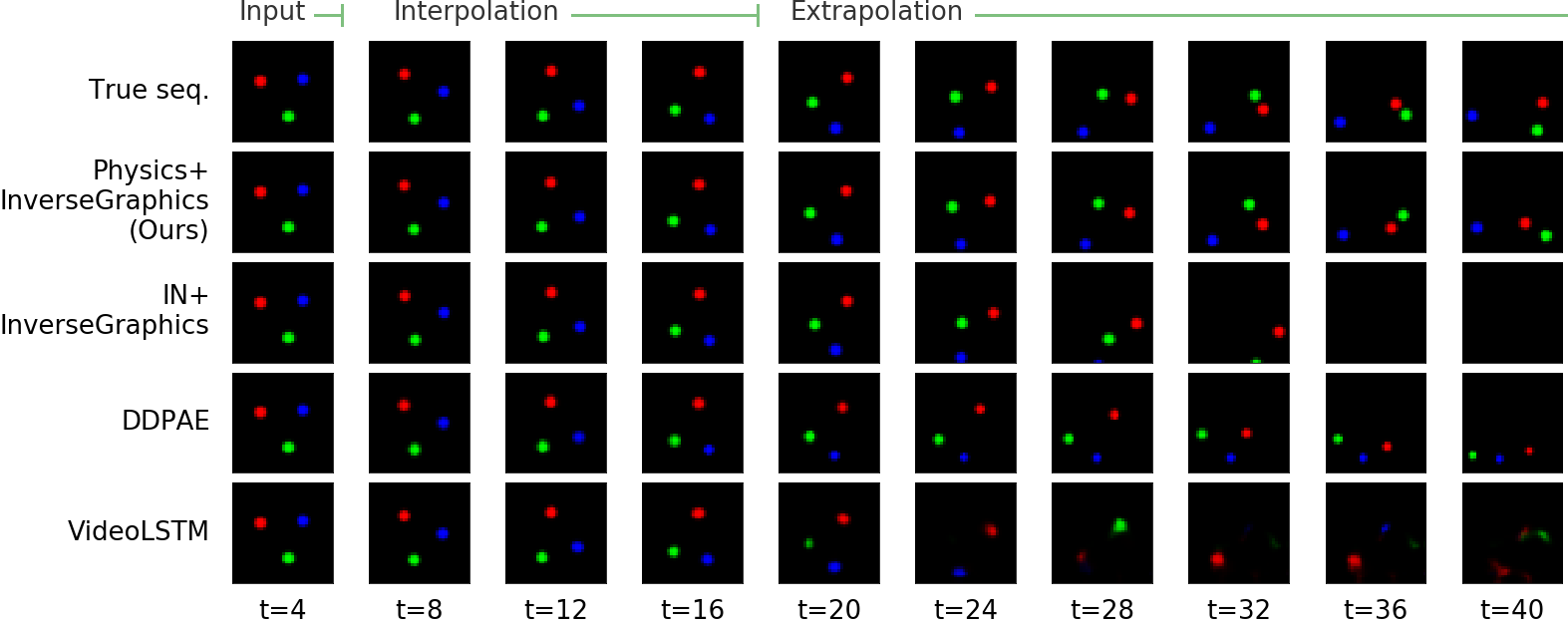}
    
    \vspace{0.3cm}
    
    \includegraphics[width=0.80\textwidth]{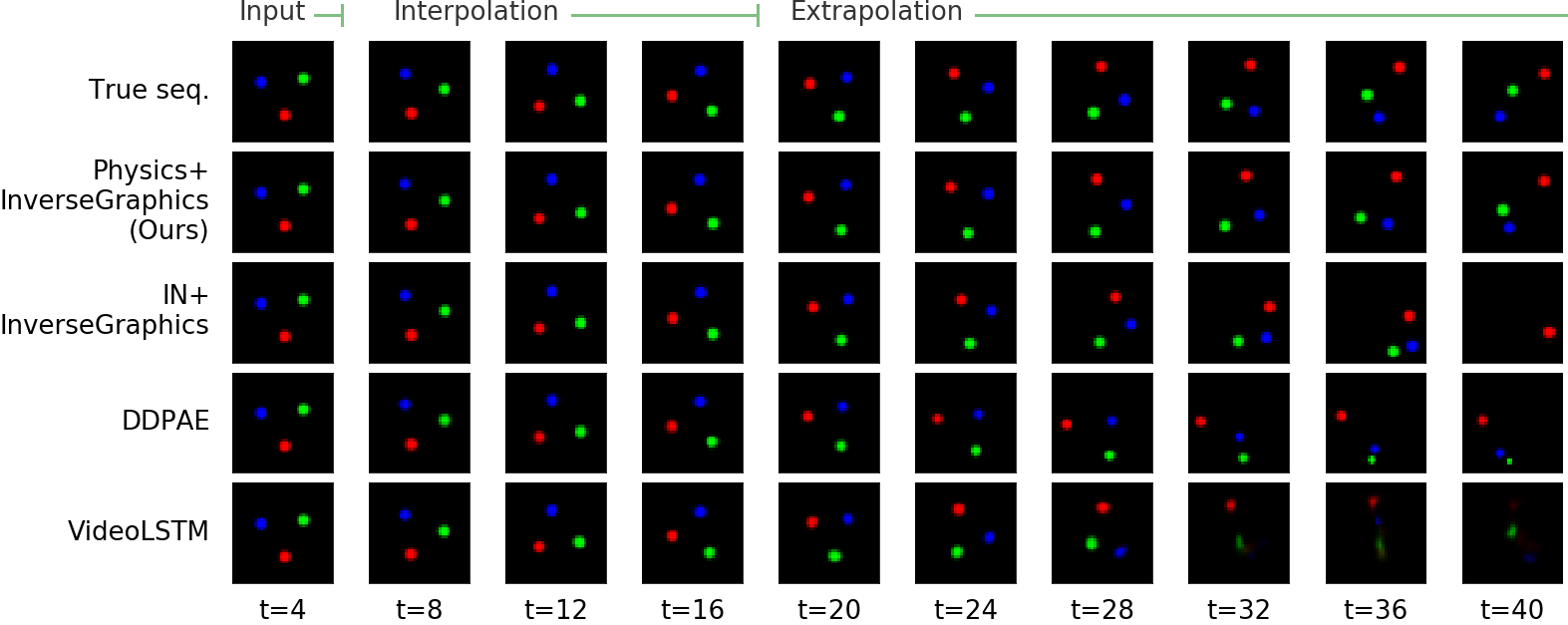}
\end{figure}

\subsubsection*{2-balls spring}
\begin{figure}[H]
  \centering
    \includegraphics[width=0.80\textwidth]{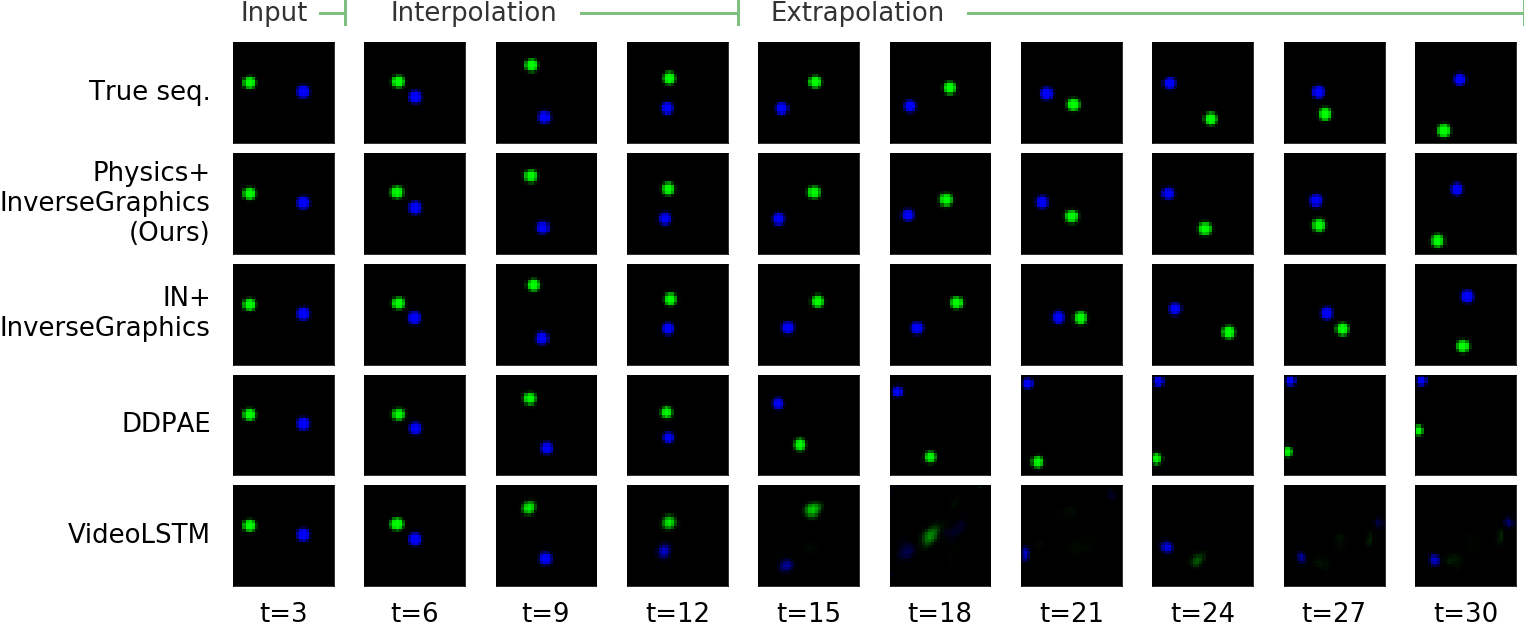}
    
    \vspace{0.3cm}
    
    \includegraphics[width=0.80\textwidth]{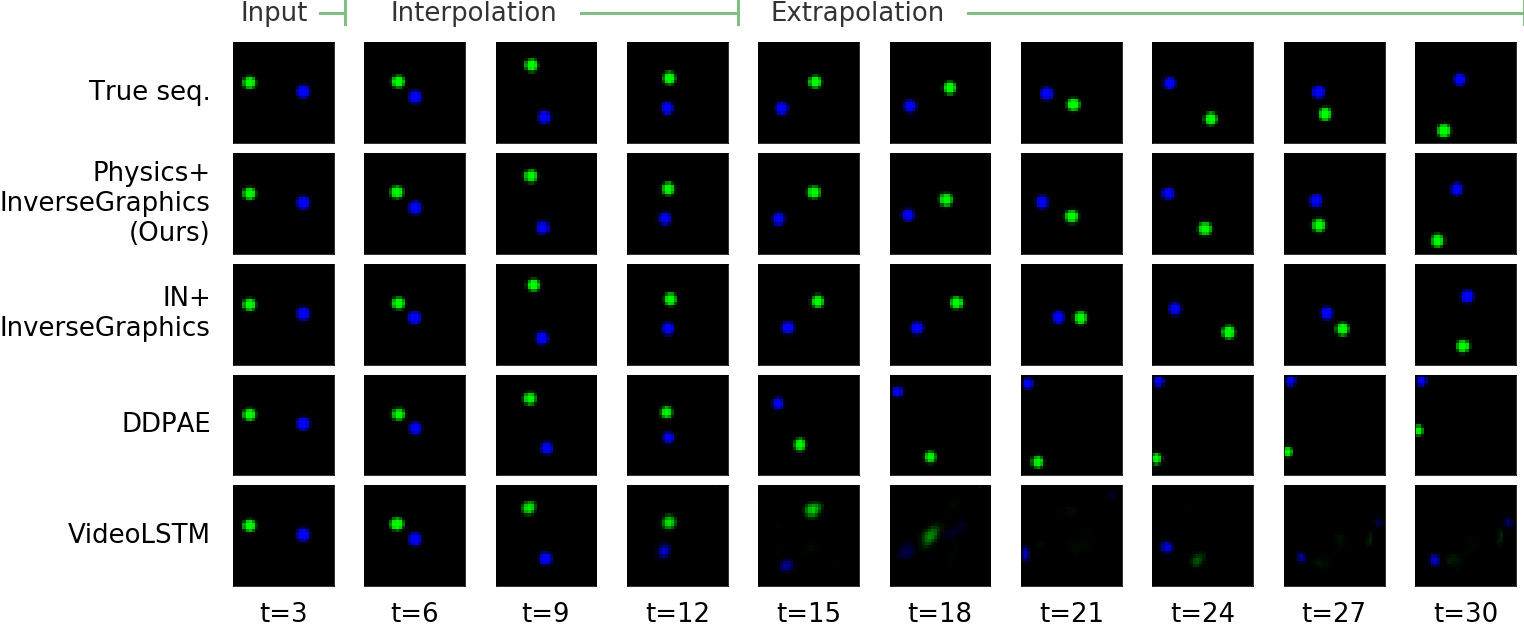}
\end{figure}

\pagebreak

\subsubsection*{2-balls bouncing}
\begin{figure}[H]
  \centering
    \includegraphics[width=0.80\textwidth]{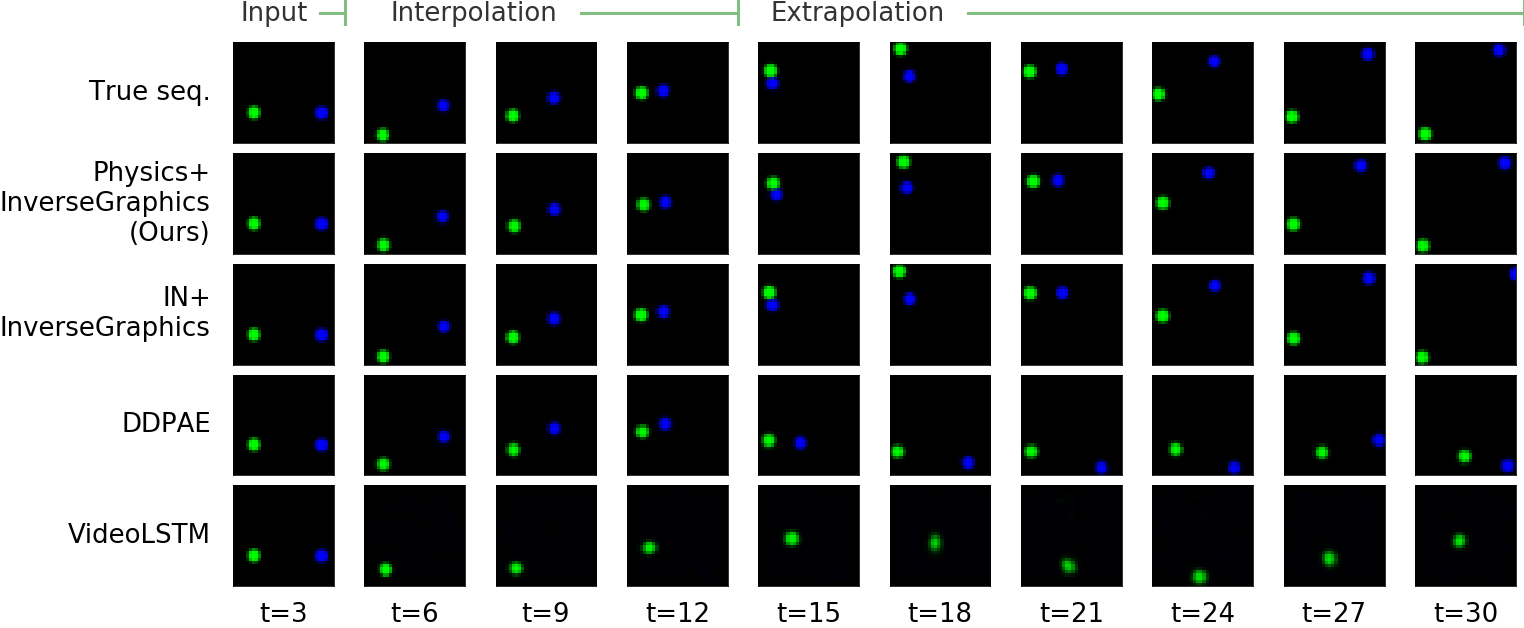}
    
    \vspace{0.3cm}
    
    \includegraphics[width=0.80\textwidth]{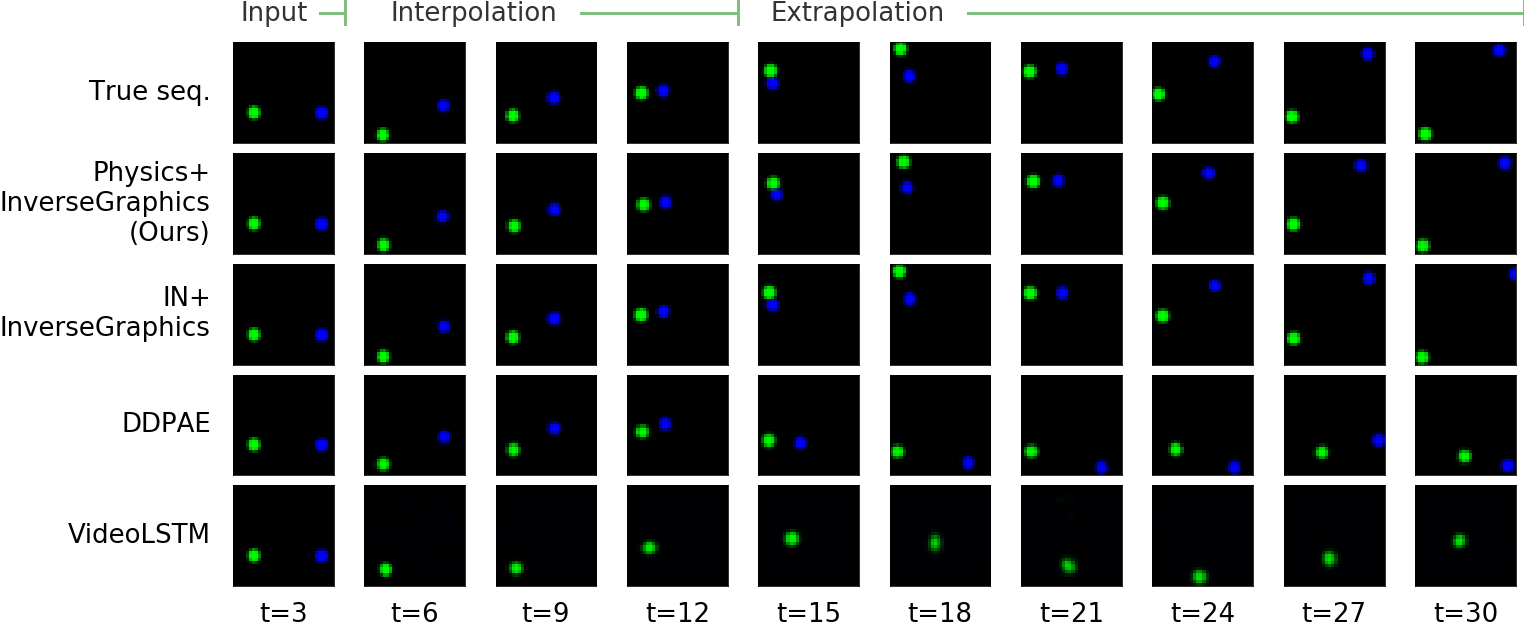}
\end{figure}

\subsubsection*{2-digits spring}
\begin{figure}[H]
  \centering
    \includegraphics[width=0.80\textwidth]{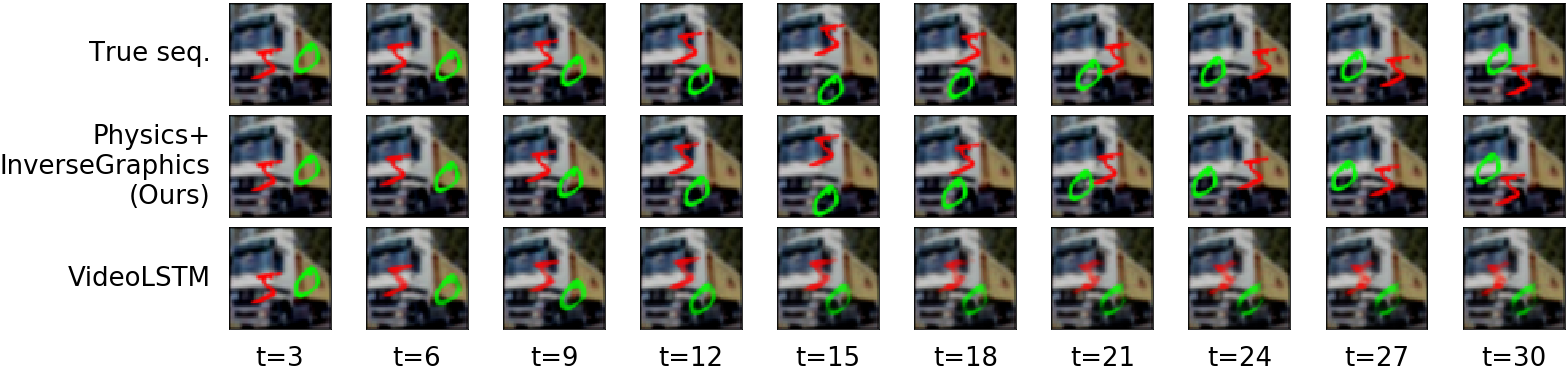}
    
    \vspace{0.3cm}
    
    \includegraphics[width=0.80\textwidth]{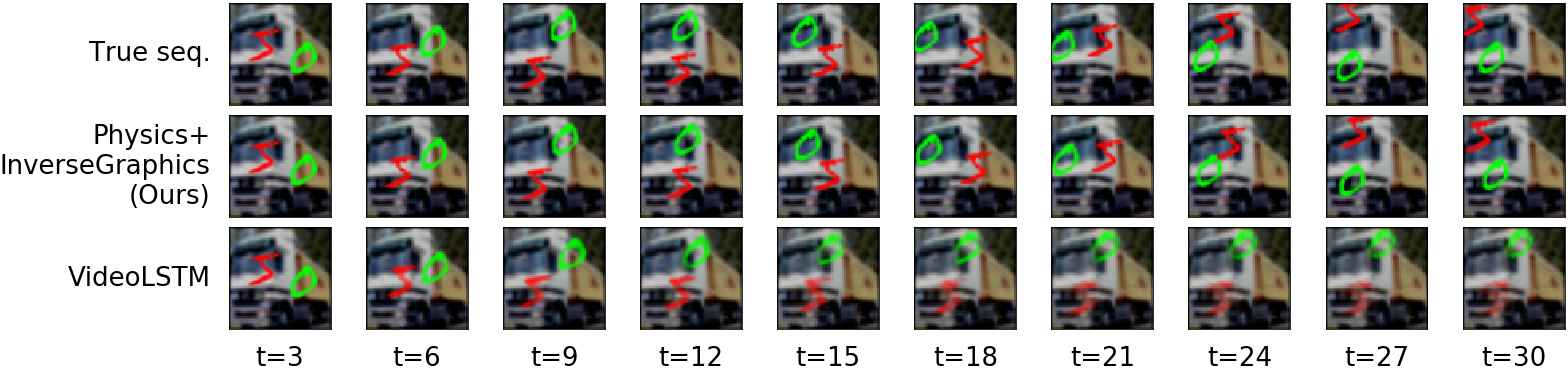}
\end{figure}

\section{Extrapolation to unseen image regions}

One limitation of standard fully-connected or deconvolutional decoders is the inability to decode states corresponding to object poses or locations not seen during training. For example, if in the training set no objects appear in the bottom half of the image, a fully-connected decoder will simply learn to output zeros in that region. If in the test set objects move into the bottom half of the image, the decoder lacks the inductive bias necessary to correctly extrapolate in image space.

\begin{figure}[h]
  \centering
    \includegraphics[width=0.8\textwidth]{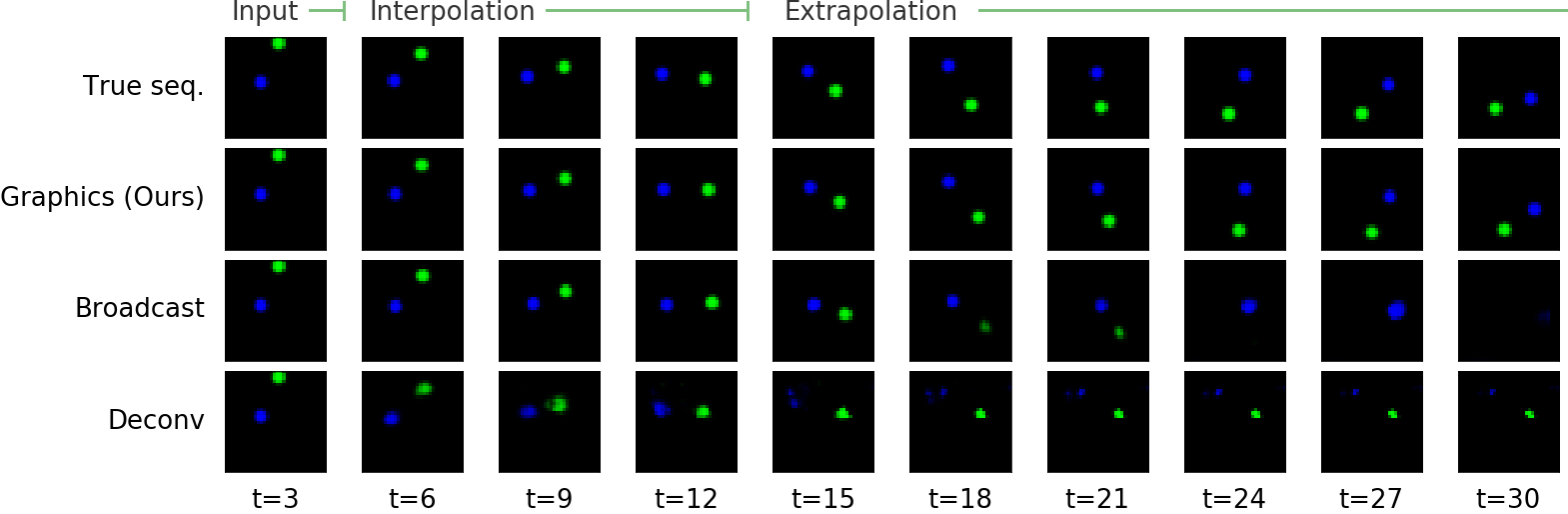}
    \label{fig:decoder_ablations}
    \caption{Comparison between graphics decoder and two black-box decoders, trained on data where objects only appear in the top half of the scene. Only the graphics decoder is able to correctly render the objects in the  bottom half of the scene at test time. Broadcast: spatial broadcast decoder \citep{Watters2019SpatialVAEs}; Deconv: standard deconvolutional network.}
\end{figure}

To test this hypothesis, we replaced our model's decoder with a Deconv and Spatial Broadcast \citep{Watters2019SpatialVAEs} decoder, and compared them in a spatial extrapolation experiment. In this experiments, objects never enter the bottom half of the image in the input and prediction range, though in the extrapolation range in the test set objects move to this region of the scene.
In the rollouts shown in Fig. \ref{fig:decoder_ablations}, Broadcast performs better than Deconv, but they both fail to maintain object integrity when the balls
move to the bottom half of the image in the extrapolation steps, validating our hypothesis that a black-box decoder has insufficient extrapolation ability. In contrast, our rendering decoder is be able to correctly decode states not seen during training. 

In the limit that our renderer corresponds to a full-blown graphics-engine, any pose, location, color, etc. not seen during training can still be rendered correctly. This property gives models using rendering decoders, such as ours and \cite{Hsieh2018LearningPrediction}, an important advantage in terms of data-efficiency. We note, however, that in general this advantage does not apply to correctly inferring the states from images whose objects are located in regions not seen during training. This is because the encoders used are typically composed simply of convolutional and fully-connected layers, with limited de-rendering inductive biases.

\section{Incorrect number of object slots}
\begin{figure}[h]
  \centering
    \includegraphics[width=0.21\textwidth]{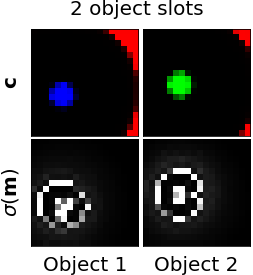}
    \hspace{1cm}
    \includegraphics[width=0.4\textwidth]{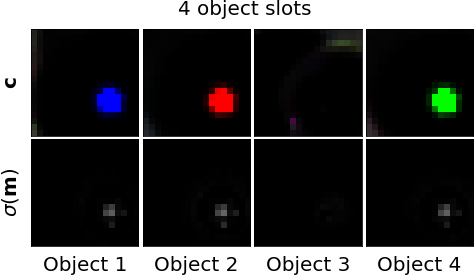}
    \caption{Results for incorrect number of object slots in the physics engien for the 3-body gravitational system \textbf{Left:} Contents and masks learned for 2 object slots. \textbf{Right:} Contents and objects learned for 4 object slots.    \label{fig:slots}
}
\end{figure}

The model proposed assumes we know the number of objects present in the scene. Here we briefly explore how to the model behaves when we use an incorrect number of slots $N$. We use the gravitational system, since interaction forces between objects are easy to generalize for any $N$. Fig. \ref{fig:slots}, left, shows that when using only 2 object slots, two of the objects are found, since the model does not have capacity to find more. Fig. \ref{fig:slots}, right, shows that when using more slots than the number of objects in the scene, all objects are discovered, and extra slots are left empty. However, in both cases we found predictive performance to be subpar, since in one case there are objects missing to correctly infer interactions, and in the other there are interactions between object slots and empty slots, confusing the dynamics. 

\end{document}